\documentclass[10pt]{article}

\newif\ifanon
\anonfalse
\ifanon
\usepackage{tmlr}
\else
\usepackage[preprint]{tmlr}
\fi

\usepackage[utf8]{inputenc}
\usepackage{amsmath,amssymb,amsthm}
\usepackage{booktabs}
\usepackage{graphicx}
\usepackage{hyperref}
\usepackage{url}
\usepackage[capitalize,noabbrev]{cleveref}
\usepackage{pgfplots}
\pgfplotsset{compat=1.18}
\usepackage{xcolor}

\hypersetup{
    colorlinks=true,
    linkcolor=blue,
    citecolor=blue,
    urlcolor=blue
}

\crefname{appendix}{Appendix}{Appendices}
\Crefname{appendix}{Appendix}{Appendices}

\newtheorem{proposition}{Proposition}

\title{Spectral Margins and Memory Routing\\in Looped Transformers}

\author{\name Matthias M.\ M.\ Buehlmaier \email buehl@hku.hk \\
        \addr Faculty of Business and Economics, The University of Hong Kong}

\begin{document}

\maketitle

\begin{abstract}
Looped transformers apply one block of weights repeatedly. Each iteration keeps a fraction of the previous hidden state, the \emph{carry}, and adds the block's output, so information can pass between iterations through either path. We study a small, LayerNorm-based variant of the Parcae architecture and ask two questions: does the iterated state settle, and does learning to remember across iterations require the carry? On stability, pre-LayerNorm makes the block less sensitive at its fixed point as the carry grows, but not enough to guarantee contraction: with a scalar carry, a perturbation can grow in a single iteration even at a locally stable fixed point. These fixed points are stable because the spectral radius of the loop's Jacobian stays below $1$; its distance from $1$, the spectral margin, narrows as the carry approaches $1$ and, at high carry, as the width grows, and some initializations do not settle within the iteration budget. On memory, a multi-hop copy task trained at a fixed loop depth is learned as well with the carry frozen at $0.5$, $0.02$, or $0$ as with a learned carry, which changes little during training. When the depth is sampled during training, a carry of $0.5$ still trains but $0.02$ and $0$ do not. All experiments are small ($d \le 32$ for training).
\end{abstract}

\section{Introduction}

Looped transformers reuse a single block of transformer weights across depth: a model with $k$ unique layers iterated $T$ times has the effective depth of a $kT$-layer network while storing only the $k$ layers. We study the recurrence $h_{t+1} = \bar{A} \odot h_t + f(h_t, e)$ (\cref{sec:setup}; \cref{fig:two-paths}). Each iteration adds two terms: a retained fraction of the previous hidden state (the \emph{carry} $\bar{A} \odot h_t$, a per-channel linear path) and an update computed from that state and the re-injected input (the block \emph{increment} $f$, a dense nonlinear path). The carry is Parcae's diagonal state-transition matrix~\citep{parcae2026}, what state-space models call the state matrix, gated RNNs a forget gate, and Highway networks a carry gate~\citep{srivastava2015}. Since both terms depend on the previous state, information can pass between iterations through either path.

This raises two questions. Does the hidden state settle as iterations accumulate, and what guarantees that it does? And when a task requires remembering information across iterations, does learning it require the carry? Both questions concern the carry, which plays two roles: it shapes the loop's dynamics, and it is a candidate path for the memory a finite-depth computation needs. Parcae's stability guarantee rests on constraining it. Our results keep the two roles apart: boundedness does not imply convergence, convergence does not imply a useful answer, and at fixed depth, learning a useful computation does not require a linear carry. The clearest instance is a trained model whose fixed point is more stable than at initialization, while the correct answer is available only in a transient a few iterations long (\cref{subsec:intervention}).

We study a small Parcae-style model with pre-LayerNorm blocks (Parcae uses RMSNorm), without Parcae's input-injection matrix and prelude normalization, and trained at a fixed loop depth except in \cref{subsec:sampled-depth}; \cref{sec:setup} details the differences.

\begin{figure}[t]
\centering
\begin{tikzpicture}[
    box/.style={draw, minimum width=2.2cm, minimum height=0.8cm, align=center, font=\small},
    arr/.style={->, thick},
    sum/.style={draw, circle, minimum size=0.6cm, inner sep=0pt},
    io/.style={font=\small},
    note/.style={font=\footnotesize, text=gray!60!black, align=center},
]
\node[io] (h) at (0, 0.35) {$h_t$};
\node[io] (e) at (0, -1.15) {$e$};
\node[box, fill=blue!15] (carry) at (3.2, 0.7) {$\bar{A} \odot h_t$};
\node[box, fill=green!15] (block) at (3.2, -0.8) {$f(h_t, e)$};
\node[sum] (sum) at (5.9, -0.05) {$+$};
\node[io] (out) at (7.3, -0.05) {$h_{t+1}$};
\draw[arr] (h) -- (carry);
\draw[arr] (h) -- (block);
\draw[arr] (e) -- (block);
\draw[arr] (carry) -- (sum);
\draw[arr] (block) -- (sum);
\draw[arr] (sum) -- (out);
\node[note] at (3.2, 1.55) {linear path (the carry):\\diagonal, per-channel rescaling};
\node[note] at (3.2, -1.75) {nonlinear path (the block increment):\\dense: attention $+$ MLP, pre-LN};
\end{tikzpicture}
\caption{Each loop iteration sums two paths into the next state: the \emph{carry} retains a per-channel fraction $\bar{A}_i \in (0,1)$ of the previous state (linear and diagonal: it mixes neither channels nor positions), and the block \emph{increment} $f$ computes a dense nonlinear update from the state and the re-injected input $e$. Does their sum settle, and does learning to remember require the carry?}
\label{fig:two-paths}
\end{figure}

The contributions are the following, for looped transformers with pre-LayerNorm blocks and a Parcae-style diagonal carry.
\begin{itemize}
    \item \textbf{Gain attenuation without contraction} (\cref{sec:lipschitz,sec:self-stabilization}). LayerNorm couples the block's local Lipschitz constant inversely to the activation scale, and at a fixed point an exact identity writes it as $1 - \rho$ times a ratio that we measure to be nearly constant across scalar carries (\cref{prop:homogeneity,prop:balance}). Yet with scalar carry, every verified fixed point has a recurrence Jacobian with operator norm above $1$ and spectral radius below: it is strongly non-normal.
    \item \textbf{The spectral margin and its exact structure} (\cref{subsec:directional,sec:per-channel}). The stability margin $1 - \rho_{\text{spec}}(J_G)$ shrinks as $\rho \to 1$ and, at $\rho = 0.9$, roughly as $d^{-2/3}$ with width up to $d = 256$, while a minority of initializations do not settle within $2 \times 10^4$ iterations. Two exact facts organize the spectrum: the carry $\rho$ is itself an eigenvalue (\cref{prop:mean-shift}), and causal attention makes the spectrum a per-token property (\cref{prop:block-triangular}). In the per-channel profiles tested, a single slow channel narrows the margin to near zero.
    \item \textbf{At fixed depth, the carry is not needed to learn the memory task} (\cref{subsec:windowed,subsec:ablation,subsec:intervention}). On a copy task that forces information through several iterations, a carry frozen at $0.5$, $0.02$, or $0$ trains as well as a learned one, and training leaves a learned carry near its initialization (five of five seeds; nine of ten for the zero carry, the tenth an optimization failure). The block is necessary for the relay. A model trained with a learned carry nonetheless breaks if the carry is zeroed afterwards, so it depends on the carry at run time, in a way we do not resolve. The trained model answers correctly only within a few iterations of its training depth.
    \item \textbf{Under sampled depth, the carry's size matters} (\cref{subsec:sampled-depth}). With the depth drawn per batch, as in Parcae's training, a carry of $0.5$, learned or frozen, trains, while $0.02$ and $0$ stay at chance. A constructed per-channel task shows the converse regime, in which training moves the carry to track the task's per-channel rates (\cref{subsec:boundary}).
\end{itemize}

The exact results are dimension-independent. Everything else is measured at small scale ($d = 16$--$32$ for training, up to $256$ for random-weight spectra); \cref{sec:limitations} collects the scope conditions and \cref{sec:reproducibility} the implementation.

\section{Related Work}
\label{sec:related}

\textbf{Parcae}~\citep{parcae2026} identifies two stability mechanisms, a spectral-norm constraint on the diagonal state-transition matrix (the carry) and prelude normalization. We add the in-block normalizer's gain control and show that, on the tasks tested, the carry is not needed to learn depth memory at fixed depth. Parcae also fit a saturating-exponential test-time scaling law $\mathcal{L}(T) = \mathcal{L}_\infty + Ze^{-zT}$; \cref{subsec:directional} offers a candidate mechanism.

\textbf{S4/Mamba}~\citep{s4,mamba}. The constrained discretization Parcae imports from S4, which builds on HiPPO~\citep{hippo} and was later reduced to diagonal and scalar state matrices~\citep{s4d,mamba2}, guarantees $\rho < 1$. Our result concerns the nonlinear dependence on the state that S4's linear state recurrence lacks; the per-channel analysis (\cref{sec:per-channel}) shows the memory--stability tradeoff of the diagonal state matrix transfers intact to the depth axis. The linear recurrent unit of \citet{orvieto2023} uses the same exponential parameterization for the magnitudes of its complex diagonal recurrence, with eigenvalues deliberately placed near the unit circle for long memory.

\textbf{Normalization and recurrent dynamics.} That normalization stabilizes recurrent hidden states dates to LayerNorm's introduction~\citep{layernorm}, and the inverse scaling of a scale-invariant map's Jacobian is a known consequence of homogeneity~\citep{arora2019}, exploited for pre-LN training~\citep{xiong2020}. Signal-propagation analyses study how attention and normalization condition the forward map across the layers of a deep transformer~\citep{dong2021,noci2022}; here depth is a loop of one weight-tied block and the object of study is its fixed point. Closest to this work, \citet{tomihari2025} study recurrent self-attention through state Jacobians, finding that normalization suppresses Lipschitz constants and oscillatory components and places the dynamics near criticality; the width collapse of the spectral margin in \cref{subsec:directional} is a measurement of that near-criticality on the depth axis. Our contribution is specific to the gated carry: the fixed-point balance that yields $\text{Lip}(f) \propto 1 - \rho$, the operator-norm/spectral-radius decomposition at verified fixed points, the exact LayerNorm eigenmodes, and the training-side question of where memory is routed. \citet{kim2021} show that standard dot-product self-attention is not Lipschitz on an unbounded domain and bound the constant of an $L_2$ variant; the Lipschitz statements here are local, at bounded operating points of the standard block.

\textbf{Non-normal recurrences.} Operator norm above $1$ with spectral radius below is the signature of non-normality, whose dynamical content is transient growth~\citep{trefethen2005}; \citet{kerg2019} and \citet{orhan2020} exploit it deliberately in recurrent networks. \citet{miller2019} show that recurrent models whose state map is contractive can be approximated by feed-forward ones; the looped blocks studied here are not contractive in operator norm at any tested scalar carry. Over a finite loop depth $T$ the relevant quantity is the transient amplification along the trajectory rather than the asymptotic rate, a point \cref{subsec:directional} returns to.

\textbf{Looped transformers}~\citep{dehghani2019,giannou2023,yang2024,saunshi2025,geiping2025,kohli2026} motivate looping to greater depth at test time; our analysis characterizes the dynamical regime that keeps such looping well-behaved. Weight-tied recurrent networks for algorithmic extrapolation~\citep{schwarzschild2021,bansal2022} likewise iterate one block beyond the training depth; \citet{bansal2022} keep an explicit copy of the input available to the recurrence and name the degradation under excess iterations \emph{overthinking}.

\textbf{Deep equilibrium models}~\citep{deq} solve for the fixed point of a weight-tied block by root-finding and differentiate implicitly at equilibrium, with follow-up work on guaranteed convergence~\citep{monotonedeq} and Jacobian regularization~\citep{bai2021}. The looped transformers here train through a fixed finite depth, and the trained model of \cref{subsec:intervention} computes in a transient and loses its answer at its own fixed point, the regime a DEQ computes in by construction. The fixed-point machinery nevertheless transfers.

\section{Setup and Notation}
\label{sec:setup}

Consider the Parcae recurrence for a looped transformer:
\begin{equation}
h_{t+1} = \bar{A} \odot h_t + f(h_t, e)
\label{eq:recurrence}
\end{equation}
where
\begin{itemize}
    \item $h_t \in \mathbb{R}^{n \times d}$ is the hidden state at loop iteration $t$ (sequence length $\times$ embedding dimension);
    \item $\bar{A} \in (0,1)^d$ is the constrained diagonal carry and $\odot$ is the elementwise (Hadamard) product, broadcast over positions: $(\bar{A} \odot h_t)_{ji} = \bar{A}_i\,(h_t)_{ji}$, equivalently $h_t\,\mathrm{diag}(\bar{A})$. Channel $i$ retains the fraction $\bar{A}_i$ of its previous state every iteration (a \emph{channel} is one coordinate of the representation, at every position). The scalar case $\bar{A} = \rho\mathbf{1}$ is the warm-up; in general each channel has its own rate. Parcae's parameterization $\bar{A} = \exp(\Delta t \cdot (-\exp(\lambda)))$ with learnable $\lambda \in \mathbb{R}^d$ (Parcae's $\log A$) keeps every channel in $(0,1)$ by construction while training moves it;
    \item $e$ is the encoded input from the prelude (the embedding stage before the loop), re-injected each iteration;
    \item $f$ is the transformer block \emph{increment}, the block's output minus its input, $f(h_t,e) = \text{Block}(h_t + e) - (h_t + e)$, comprising pre-LN multi-head self-attention and a pre-LN MLP. The block's internal residual path is carried by $\bar{A} \odot h_t$, so $f$ contains no identity in $h_t$ (\cref{app:coupling}). We also call $f$ the \emph{nonlinear path} when contrasting it with the linear carry.
\end{itemize}

We write $\rho = \|\bar{A}\|_\infty$ for the carry's spectral radius (its largest entry). The spectral radius of the full recurrence Jacobian, a different and generally larger quantity, is always written $\rho_{\text{spec}}(J_G)$. We write $G(h) = \bar{A} \odot h + f(h, e)$ for the loop map, and call a fixed point \emph{verified} when its relative residual $\|G(h) - h\|/\|h\|$ is below $10^{-9}$. The \emph{readout} is the stage after the loop: a final LayerNorm, then the unembedding projection.

\paragraph{Relationship to Parcae.} The default implementation is a Parcae-inspired gated recurrent-depth transformer, not a faithful reproduction: no explicit $\bar{B} \cdot e$ injection (input enters only through the block), no prelude normalization, $\Delta t = 1$ fixed, and a fixed loop depth except in the sampled-depth experiment of \cref{subsec:sampled-depth}. The omitted terms are functions of the input $e$, which is constant across the loop, so they contribute no direct Jacobian in $h_t$ and leave the form of the local analysis unchanged; they can shift the operating point, so every measured value is specific to this variant. A variant restoring $\bar{B} \cdot e$, learnable $\Delta$, and prelude normalization reproduces the central windowed-copy result (every stage solved, carry not recruited). Parcae's trained models use RMSNorm and stochastic depth sampling and operate at vastly larger scale.

\textbf{Norms.} $\|\cdot\|$ on states is the Frobenius norm over positions and channels; $h$ denotes the state and $H$ its $n \times d$ matrix form. We write $\text{Lip}(f)|_h := \|J_f(h)\|_2$ for the local Lipschitz constant of the increment at $h$, and $\text{Lip}(f)$ when the point is clear. LayerNorm statistics $\mu, \sigma$ are per token; the diagonal carry scales channel $i$ by $\bar{A}_i$ at every position.

\paragraph{Terminology.} We keep three progressions apart: the \emph{position} along the sequence; the \emph{loop iteration} $t$, along which the state evolves while the input and the weights stay fixed; and the \emph{training step}, at which the weights, including the carry, change. We likewise distinguish whether trajectories stay \emph{bounded}, whether a fixed point is \emph{locally stable}, and whether the model performs its task at a \emph{finite depth} $T$. A model can read out a useful answer after $T$ iterations without having reached a fixed point, and the trained model of \cref{subsec:intervention} does exactly that. The carry is \emph{recruited} when training moves it substantially in the direction the task's memory demand predicts.

\paragraph{A scalar warm-up.} If the update were a constant $c$, the recurrence $h_{t+1} = a\,h_t + c$ with $0 < a < 1$ would settle at $h^* = c/(1-a)$: with $a = 0.5$ and $c = 2$ the iterates run $0, 2, 3, 3.5, \ldots \to 4$, and with $a = 0.9$ the same update settles at $20$. Retaining more of the previous state supports a proportionally larger settled state. The transformer's increment depends on $h_t$, so settling is no longer automatic, but the balance equation survives as \cref{eq:fixed-point}. The stability question is also visible in one dimension: near a settled value, a disturbance evolves as $\delta_{t+1} \approx (a + f'(h^*))\,\delta_t$, so it dies out exactly when $|a + f'(h^*)| < 1$. \Cref{sec:self-stabilization} is the matrix version of this, $J_G = \text{diag}(\bar{A}) + \partial f/\partial h$, with one new phenomenon: in more than one dimension ``the size of $J_G$'' splits into the operator norm and the spectral radius, which coincide in the scalar case and are measured here to sit far apart.

Parcae's stability argument is that $\rho < 1$ by construction bounds the linear part, with the nonlinear block treated as a perturbation that prelude normalization and regularization keep in check. We show that pre-LayerNorm does more: it couples the block to the activation scale and thereby to the carry.

Three qualitatively different things can happen when \cref{eq:recurrence} is iterated, and all three appear in this paper (\cref{fig:three-fates}): the step size can decay geometrically (the state \emph{settles}); it can neither decay nor grow (a \emph{bounded, non-settling orbit}); or it can grow without bound (\emph{divergence}). Settling is the regime the fixed-point analysis addresses and Parcae's test-time scaling exploits. Bounded non-settling occurs for a minority of random initializations (\cref{tab:seed-robustness}); whether it harms a finite-depth computation depends on the readout. Divergence is what the carry constraint prevents, and what an unconstrained carry learns under aggressive training (\cref{subsec:unconstrained}).

\section{The Lipschitz Structure of LayerNorm}
\label{sec:lipschitz}

LayerNorm maps $x \mapsto \gamma \odot \frac{x - \mu}{\sigma} + \beta$ with per-token statistics $\mu, \sigma$. Two inputs show what matters. First, $\text{LN}([1,2,3])$ and $\text{LN}([10,20,30])$ are identical at $\varepsilon = 0$, and equal up to $\varepsilon$'s effect otherwise: the output is pinned regardless of the input's scale. Second, the governing variable is the \emph{spread}, not the magnitude: $[100, 100, 100]$ has a large norm but zero spread and sits at the normalizer's singularity (capped only by $\varepsilon$), while a small-norm token with healthy spread is deep in the well-behaved regime.

The first observation has a differential consequence: a function that ignores scale has a derivative that scales inversely. For one token write $P = I - \tfrac{1}{d}\mathbf{1}\mathbf{1}^\top$ for the centering projection and $s(x) = \sqrt{\tfrac{1}{d}\|Px\|^2 + \varepsilon}$ for the spread.

\begin{proposition}[Inverse-scale Jacobian]
\label{prop:homogeneity}
Let $N$ be a scale-invariant normalizer, $N(\alpha x) = N(x)$ for all $\alpha > 0$ (LayerNorm, RMSNorm, and ScaleNorm at $\varepsilon = 0$). Wherever $N$ is differentiable, $J_N(\alpha x) = J_N(x)/\alpha$ and $J_N(x)\,x = 0$. For LayerNorm with any $\varepsilon \ge 0$, moreover, $J_{\mathrm{LN}}(x)\,\mathbf{1} = 0$ and $\|J_{\mathrm{LN}}(x)\|_2 \le \|\gamma\|_\infty / s(x)$.
\end{proposition}

\noindent The proof is in \cref{app:coupling}. For a pre-LN block the increment inherits the bound: its local Lipschitz constant scales inversely with the per-token spread, $\text{Lip}(f) \propto 1/s$. Three qualifications apply. \emph{(i)} LayerNorm acts per token, so the relevant scale is the \emph{smallest} per-token spread; a global norm $\|H\|$ is a proxy that presumes token spreads scale together. \emph{(ii)} \Cref{prop:homogeneity} concerns each normalizer, not the composed block: the increment chains two normalized sublayers with an internal residual, so $f$ is not exactly scale-invariant, and the inverse coupling is the leading behavior where the state dominates the second normalizer's input (\cref{app:coupling}). \emph{(iii)} $s \le \text{RMS}$ per token, with equality when the per-token mean is small and exactly for RMSNorm.

\Cref{fig:implosion-law} shows the coupling measured on the full block with true operator norms (top singular value of the materialized increment Jacobian). Starting from a verified fixed point, we rescale the normalizer input by factors down to $3 \times 10^{-3}$ at two carries: $\text{Lip}(f)$ ranges from $0.24$ to $981$ while the product $\text{Lip}(f) \cdot \min_{\text{token}} s$ stays between $2.2$ and $4.2$ across three and a half decades. The same figure shows that when only the state is rescaled and the injection $e$ keeps its natural scale, the normalizer input cannot implode below the injection's own spread ($\approx 0.8$), and $\text{Lip}(f)$ saturates near $4$. In isolation the inverse scaling is expected, since LayerNorm divides by the standard deviation; \cref{sec:self-stabilization} turns to how it interacts with the carry.

\begin{figure}[ht]
\centering
\begin{tikzpicture}
\begin{loglogaxis}[
    width=0.72\textwidth,
    height=0.5\textwidth,
    xlabel={smallest per-token spread $\min_{\text{token}} s(h + e)$ at the normalizer input},
    ylabel={$\text{Lip}(f) = \|J_f\|_2$},
    xmin=0.002, xmax=20,
    ymin=0.1, ymax=2000,
    legend style={at={(0.5,-0.22)}, anchor=north, font=\footnotesize, fill opacity=0.9, text opacity=1},
    legend columns=2,
    grid=major, grid style={gray!25},
    tick label style={font=\footnotesize},
    label style={font=\footnotesize},
]
\addplot[gray, dashed, thick, domain=0.002:20, samples=2] {3/x};
\addlegendentry{$\text{Lip}(f) = 3/s$ (slope $-1$)}
\addplot[blue, thick, mark=*, mark size=1.8pt] coordinates {
    (1.3417,1.6277) (0.4025,7.3847) (0.1342,28.1451) (0.0403,102.5126) (0.0134,311.7791) (0.004,981.0317)
};
\addlegendentry{$\rho = 0.3$, state and injection rescaled}
\addplot[red!80!black, thick, mark=square*, mark size=1.8pt] coordinates {
    (11.5083,0.2391) (3.4525,0.7518) (1.1508,2.364) (0.3453,9.9723) (0.1151,33.4626) (0.0345,115.651)
};
\addlegendentry{$\rho = 0.9$, state and injection rescaled}
\addplot[green!50!black, thick, densely dotted, mark=triangle*, mark size=2.2pt] coordinates {
    (11.5083,0.2391) (3.4792,0.7459) (1.3549,2.098) (0.8933,3.617) (0.8308,3.9661) (0.8232,4.0401)
};
\addlegendentry{$\rho = 0.9$, state only rescaled (injection fixed)}
\draw[<-, green!50!black, thick] (axis cs:0.78,3.6) -- (axis cs:0.06,0.7) node[left, font=\footnotesize, text=green!50!black, align=center] {injection floor:\\spread $\approx 0.8$, $\text{Lip}(f) \approx 4$};
\end{loglogaxis}
\end{tikzpicture}
\caption{The gain-control law on the full block ($d=32$, seq $16$, operator norms by SVD). From a verified fixed point the state (in one mode with the injection) is rescaled by factors down to $3 \times 10^{-3}$. $\text{Lip}(f)$ follows $\approx 3/s$ over three and a half decades of the smallest per-token spread $s$; with the injection at its natural scale (triangles) $\text{Lip}(f)$ saturates near $4$.}
\label{fig:implosion-law}
\end{figure}

\paragraph{Activation implosion.} The coupling makes small spread the risky direction. With state and injection imploded together, $\text{Lip}(f)$ reaches $\approx 980$ at a per-token spread of $0.004$, and perturbations injected there are amplified up to $\times 147$ over the next eight iterations; with fixed weights, every perturbed trajectory we tested nonetheless re-converges to its original fixed point. The exposure is the high-gain transient, which the backward pass traverses during training, and the practical monitor is the smallest per-token spread at the normalizer inputs, which the global $\|h\|$ only proxies (\cref{sec:implosion}).

\section{Gain Attenuation and Local Stability}
\label{sec:self-stabilization}

This section shows three things. LayerNorm's coupling shrinks the one-step bound $\rho + \text{Lip}(f)$ toward $1$ as the carry grows, but never below it (\cref{subsec:scale,subsec:opnorm}). With scalar carry, the fixed points are nonetheless locally stable: the Jacobian is strongly non-normal, and its spectral radius stays below $1$ although its operator norm does not. That stability margin shrinks with carry and width (\cref{subsec:directional}). In the per-channel profiles tested, a single slow channel narrows it to near zero (\cref{sec:per-channel}).

\subsection{The Carry Sets the Fixed-Point Scale}
\label{subsec:scale}

Write $D = \text{diag}(\bar{A})$, $G(h) = D h + f(h, e)$ for the loop map, and define the state-weighted scale factor
\begin{equation}
q_D(h) := \frac{\|(I - D)\,h\|}{\|h\|},
\qquad 1 - \rho \;\le\; q_D(h) \;\le\; 1 - \min_i \bar{A}_i .
\label{eq:qd}
\end{equation}

\begin{proposition}[Fixed-point scale]
\label{prop:balance}
At any fixed point $h^*$ of \cref{eq:recurrence}, $f(h^*, e) = (I - D)\,h^*$. Hence, with $F := \|f(h^*, e)\|$,
\begin{equation}
\|h^*\| = \frac{F}{q_D(h^*)},
\qquad
(1 - \rho)\,\|h^*\| \;\le\; F \;\le\; \Big(1 - \min_i \bar{A}_i\Big)\,\|h^*\|,
\label{eq:fixed-point}
\end{equation}
and for scalar carry $\|h^*\| = F/(1-\rho)$ exactly.
\end{proposition}

\begin{proof}
Rearrange $h^* = D h^* + f(h^*, e)$ and take norms; $I - D$ scales channel $i$ by $1 - \bar{A}_i \in [1 - \rho,\, 1 - \min_i \bar{A}_i]$.
\end{proof}

The slowest channel governs the scale only when $h^*$ carries substantial mass in it. The measured $F$ is $O(1)$ per entry, $\approx \sqrt{nd}$ in Frobenius norm ($23$--$27$ at $nd = 512$, \cref{tab:cf}), because LayerNorm normalizes internal activations, so $\|h^*\| \propto 1/(1-\rho)$, the scalar warm-up's prediction. Existence of $h^*$ does not require contraction:

\begin{proposition}[Boundedness and existence]
\label{prop:existence}
Let $\rho < 1$ and $\varepsilon > 0$. There is a constant $B$, fixed by the weights, $\gamma$, $\beta$, and the dimensions, with $\|f(h, e)\| \le B$ for every state $h$. Every trajectory satisfies $\|h_t\| \le \rho^t \|h_0\| + B(1 - \rho^t)/(1 - \rho)$, and $G$ has a fixed point with $\|h^*\| \le B/(1-\rho)$.
\end{proposition}

\noindent The proof (\cref{app:existence}) uses that every path through the increment passes through a normalizer, whose output is bounded, and Brouwer's theorem~\citep{granas2003}. Local stability at a fixed point is a separate question, governed by $J_G(h^*)$; uniqueness is not addressed.

\subsection{The One-Step Bound Stays Above One}
\label{subsec:opnorm}

When the settled state is perturbed, can the disturbance grow on the next iteration and still die out eventually? In one dimension, no; in more than one, yes, and the split organizes this section. The shear matrix $\left(\begin{smallmatrix} 0.9 & 5 \\ 0 & 0.9 \end{smallmatrix}\right)$ amplifies a disturbance along the second coordinate about fivefold in one step (operator norm above $5$), yet every disturbance decays at the asymptotic rate $0.9$ (both eigenvalues are $0.9$; the Jordan structure contributes a polynomial factor $\sim t\,(0.9)^{t-1}$ that delays but cannot prevent decay). The \textbf{operator norm} $\|J\|_2$ is the worst one-step amplification; the \textbf{spectral radius} $\rho_{\text{spec}}(J)$ is the asymptotic per-step rate; always $\rho_{\text{spec}} \le \|J\|_2$, with equality for normal matrices ($JJ^\top = J^\top J$). A large gap is the signature of non-normality. Throughout, ``locally stable'' (or locally asymptotically stable) means $\rho_{\text{spec}}(J_G(h^*)) < 1$; ``contractive'' is reserved for operator-norm statements, which bound a single step.

Write the local coupling as $\text{Lip}(f)|_{h} = C/\|h\|$ for a coefficient $C$, and combine it with \cref{prop:balance}:
\begin{equation}
\text{Lip}(f)\big|_{h^*} = \frac{C}{\|h^*\|} = \frac{C}{F}\,q_D(h^*)
\;\xrightarrow{\text{scalar carry}}\;
\frac{C}{F}\,(1 - \rho),
\qquad
\text{Lip}(G) \le \rho + \text{Lip}(f) \approx \rho + \frac{C}{F}(1 - \rho).
\label{eq:marginal}
\end{equation}
Three steps carry different kinds of evidence. \emph{Exact:} with $C := \|J_f(h^*)\|_2\,\|h^*\|$ and $F := \|f(h^*,e)\|$, the first equality is an identity at any fixed point; by itself it predicts nothing, since $C$ and $F$ depend on the operating point. \emph{Conditional:} if $C/F$ stays bounded as $\rho \to 1$, the one-step bound $\rho + \text{Lip}(f)$ approaches $1$ in that limit, and if $C/F$ is near-constant the bound is approximately affine in $\rho$. \emph{Measured:} the top singular value of the materialized $J_f(h^*)$ gives $C/F \approx 2.5$, near-constant across $\rho \in [0.3, 0.95]$ (\cref{tab:cf}), so within the tested range the bound $\rho + 2.5\,(1-\rho)$ descends from $\approx 2$ toward $1$. Across the ten seeds of \cref{tab:seed-robustness} the per-carry mean of $C/F$ is $2.6$--$2.7$ with no trend in $\rho$.

Two consequences follow. The bound exceeds $1$ at every tested carry, so it never certifies contraction. And what the coupling delivers is \emph{gain attenuation}, the bound shrinking toward $1$. That the spectral radius of the full Jacobian stays below $1$ (\cref{subsec:directional}) is a property of these instances, not a consequence of the coupling. If $J_f = (1-\rho)K$ held exactly, $J_G = \rho I + (1-\rho)K$ would have eigenvalue $2 - \rho > 1$ whenever $K$ has an eigenvalue at $2$. For per-channel carry $q_D(h^*)$ replaces $1-\rho$; since $q_D$ can sit far above $1-\rho$ (one channel at $0.99$ with state mass spread across channels gives $q_D \approx 0.5$), the per-channel bound certifies even less (\cref{tab:per-channel}).

\begin{table}[ht]
\centering
\begin{tabular}{@{}ccccc@{}}
\toprule
$\rho$ & $\text{Lip}(f) = \|J_f\|_2$ & $F = \|f(h^*,e)\|$ & $C/F = \text{Lip}(f)/(1-\rho)$ & $C/F$, seeds 1--10 \\
\midrule
0.30 & 1.628 & 22.84 & 2.33 & $2.56 \pm 0.38$ (9) \\
0.50 & 1.233 & 22.99 & 2.47 & $2.56 \pm 0.32$ (9) \\
0.70 & 0.833 & 23.22 & 2.78 & $2.71 \pm 0.53$ (8) \\
0.90 & 0.239 & 26.66 & 2.39 & $2.64 \pm 0.55$ (8) \\
0.95 & 0.119 & 27.22 & 2.38 & $2.64 \pm 0.51$ (8) \\
\bottomrule
\end{tabular}
\caption{The gain-attenuation ratio $C/F$ at verified fixed points ($d=32$, seq $16$, random weights). Columns 2--4: seed 7, the instance plotted in \cref{fig:marginal-stability}; $\|J_f(h^*)\|_2$ is the top singular value of the materialized increment Jacobian. Last column: mean $\pm$ standard deviation over the converging seeds of \cref{tab:seed-robustness} (count in parentheses).}
\label{tab:cf}
\end{table}

\subsection{Stability Comes from the Spectral Margin}
\label{subsec:directional}

If the one-step bound exceeds $1$, why do the iterations converge? With $J_G = D + J_f$, three quantities are ordered as
\begin{equation}
\rho_{\text{spec}}(J_G) \;\le\; \|J_G\|_2 \;\le\; \rho + \|J_f\|_2 .
\label{eq:hierarchy}
\end{equation}
The right inequality is the triangle inequality. It is loose because the increment's top singular direction is not aligned with the carry, but not loose enough: at $\rho = 0.3$, $\|J_G\|_2 \approx 1.74$ against $\rho + \|J_f\|_2 \approx 1.93$. The left gap is non-normality, and it is wide enough to put the spectral radius below $1$. We measure all three by power iteration on Jacobian--vector products (central-difference JVPs of $G$, with $\rho_{\text{spec}}$ from Gelfand's formula $\|J_G^k v\|^{1/k}$) and confirm them by direct eigendecomposition and SVD of the dense Jacobian in float64. The two methods agree to $\le 0.008$ on $\rho_{\text{spec}}$ and $\le 4 \times 10^{-4}$ on the operator norm, and \cref{tab:spectral,fig:marginal-stability} report the direct values.

Three findings result (\cref{tab:seed-robustness} for the seed distribution; \cref{tab:spectral} in \cref{app:width} for the representative seed 7). First, $\rho_{\text{spec}}(J_G) < 1$ at every carry for every instance that converges to a fixed point: a small perturbation may transiently grow but eventually decays. Second, $J_G$ is markedly non-normal: $\|J_G\|_2$ exceeds $1$ in every converged instance (seed means $1.88$ at $\rho = 0.3$ and $1.07$ at $\rho = 0.9$) while $\rho_{\text{spec}}$ sits at $0.84$--$0.99$. The gap is largest at low carry and closes as $\rho \to 1$, where the near-normal diagonal carry dominates $J_G$. Third, the stability margin that matters is the spectral margin $1 - \rho_{\text{spec}}$: averaged over converging seeds it is widest at low carry ($0.159$ at $\rho = 0.3$) and collapses monotonically to $0.012$ at $\rho = 0.95$. The single instance of \cref{tab:spectral} peaks near $\rho = 0.5$ instead, within the seed spread. The spectral margin measures the local asymptotic rate; for a non-normal $J_G$ it does not measure robustness to perturbation (\cref{sec:limitations}).

\begin{figure}[ht]
\centering
\begin{tikzpicture}
\begin{axis}[
    width=0.78\textwidth,
    height=0.5\textwidth,
    xlabel={Carry $\rho$},
    ylabel={amplification factor},
    xmin=0.25, xmax=1.0,
    ymin=0.75, ymax=2.05,
    legend style={at={(0.5,-0.24)}, anchor=north, font=\footnotesize, fill opacity=0.9, text opacity=1},
    legend columns=2,
    grid=major,
    grid style={gray!30},
]
% seed band: mean +/- std of rho_spec over converging seeds (Table 4)
\addplot[blue!40, fill=blue, fill opacity=0.15, draw=none, forget plot] coordinates {
    (0.3, 0.903) (0.5, 0.942) (0.7, 0.980) (0.9, 0.983) (0.95, 0.992)
    (0.95, 0.984) (0.9, 0.965) (0.7, 0.914) (0.5, 0.850) (0.3, 0.779)
} \closedcycle;
\addplot[black, dashed, thick, sharp plot, mark=none] coordinates {(0.25, 1.0) (1.0, 1.0)};
\addlegendentry{unit threshold}
\addplot[red!70!black, thick, mark=triangle*, mark size=2pt] coordinates {
    (0.3, 1.928) (0.5, 1.733) (0.7, 1.533) (0.9, 1.139) (0.95, 1.069)
};
\addlegendentry{Bound $\rho + \mathrm{Lip}(f)$, seed 7}
\addplot[orange, thick, mark=square*, mark size=1.8pt] coordinates {
    (0.3, 1.735) (0.5, 1.481) (0.7, 1.302) (0.9, 1.041) (0.95, 1.018)
};
\addlegendentry{Operator norm $\|J_G\|_2$, seed 7}
\addplot[blue, thick, mark=*, mark size=1.8pt] coordinates {
    (0.3, 0.841) (0.5, 0.896) (0.7, 0.947) (0.9, 0.974) (0.95, 0.988)
};
\addlegendentry{Spectral radius $\rho_{\text{spec}}(J_G)$, seed mean}
\addplot[blue!60, thin, mark=o, mark size=1.6pt] coordinates {
    (0.3, 0.928) (0.5, 0.901) (0.7, 0.968) (0.9, 0.974) (0.95, 0.986)
};
\addlegendentry{$\rho_{\text{spec}}(J_G)$, seed 7}
\addlegendimage{area legend, draw=none, fill=blue, fill opacity=0.3}
\addlegendentry{seed mean $\pm$ std}
\end{axis}
\end{tikzpicture}
\caption{One-step bound, operator norm, and spectral radius of $J_G$ ($d=32$, seq $16$). The bound $\rho + \text{Lip}(f) \approx \rho + 2.5(1-\rho)$ and the operator norm both exceed $1$ except as $\rho \to 1$; the spectral radius (mean $\pm$ one standard deviation over the converging seeds of \cref{tab:seed-robustness}, seed 7 overlaid) stays below $1$, and its distance from $1$ is the stability margin.}
\label{fig:marginal-stability}
\end{figure}

\begin{table}[ht]
\centering
\begin{tabular}{@{}ccccc@{}}
\toprule
$\rho$ & $\rho_{\text{spec}}(J_G)$ (mean $\pm$ std) & margin $1-\rho_{\text{spec}}$ & with $\rho_{\text{spec}} > 1$ & non-settling \\
\midrule
0.30 & $0.841 \pm 0.062$ & 0.159 & 0/9 & 1/10 \\
0.50 & $0.896 \pm 0.046$ & 0.104 & 0/9 & 1/10 \\
0.70 & $0.947 \pm 0.033$ & 0.053 & 0/8 & 2/10 \\
0.90 & $0.974 \pm 0.009$ & 0.026 & 0/8 & 2/10 \\
0.95 & $0.988 \pm 0.004$ & 0.012 & 0/8 & 2/10 \\
\bottomrule
\end{tabular}
\caption{Seed robustness: ten random-weight initializations, power iteration at verified fixed points, confirmed by direct eigendecomposition at $\rho = 0.3$ and $0.9$. Statistics are over the converging seeds. Non-settling seeds remain on bounded, non-settling orbits after $2 \times 10^4$ iterations (relative residual $10^{-4}$--$10^{-1}$), so there is no settled state at which to evaluate the Jacobian.}
\label{tab:seed-robustness}
\end{table}

\paragraph{An exact eigenvalue from shift invariance.} Let $\mathbf{1}_s$ denote the state that is $1$ in every channel of token $s$ and $0$ elsewhere.

\begin{proposition}[Mean-shift eigenvalue]
\label{prop:mean-shift}
For the pre-LN block with LayerNorm (any $\varepsilon \ge 0$), $f(h + c\,\mathbf{1}_s, e) = f(h, e)$ for every state $h$, token $s$, and scalar $c$. Consequently, for scalar carry $\bar{A} = \rho\mathbf{1}$, $J_G(h)\,\mathbf{1}_s = \rho\,\mathbf{1}_s$ at every state, so $\rho$ is an eigenvalue of $J_G(h)$ with geometric multiplicity at least $n$, and
\begin{equation}
\rho_{\text{spec}}(J_G) \;\ge\; \rho
\qquad\text{and}\qquad
1 - \rho_{\text{spec}}(J_G) \;\le\; 1 - \rho .
\label{eq:margin-cap}
\end{equation}
\end{proposition}

\begin{proof}
With $x = h + e$, the increment is $f = a(x) + m(x + a(x))$, where $a$ and $m$ are the attention and MLP sublayers, each reading its input through a LayerNorm (\cref{app:coupling}). Adding $c\,\mathbf{1}_s$ to $x$ shifts token $s$ of both normalizer inputs by a per-token constant, which the centering removes (\cref{prop:homogeneity}), so $f$ is unchanged and $J_f(h)\,\mathbf{1}_s = 0$. Then $J_G(h)\,\mathbf{1}_s = \rho\,\mathbf{1}_s$, and the $n$ vectors $\mathbf{1}_s$ are linearly independent.
\end{proof}

On the materialized Jacobian exactly $n = 16$ eigenvalues sit within $10^{-6}$ of $\rho$ at every tested carry. The cap does not bind in the tested range: the measured $\rho_{\text{spec}}$ ($0.928$ at $\rho = 0.3$ for this instance, $0.84$ averaged over seeds) sits far above $\rho$, and at all five carries the spectral radius of the induced map on the centered quotient equals the full $\rho_{\text{spec}}$. The dominant slow mode is a centered one, so the measured margins are not artifacts of mean modes the readout's final LayerNorm would discard. For non-uniform carry $D\,\mathbf{1}_s$ is no longer proportional to $\mathbf{1}_s$ ($16$--$19\%$ of its norm lies off the mean-shift line for the profiles of \cref{tab:per-channel}), so the mean direction feeds back into the centered dynamics.

\paragraph{Causal attention makes the Jacobian block triangular.}

\begin{proposition}[Per-token spectrum]
\label{prop:block-triangular}
If attention is causal (full or windowed), then with the state ordered by token position $J_G(h)$ is block lower triangular with $d \times d$ blocks, and its spectrum is the union of the spectra of the diagonal blocks $J_{ss} = \partial G_s / \partial h_s$.
\end{proposition}

\begin{proof}
LayerNorm, the MLP, and the carry act per token, and causal attention at position $s$ reads only positions $s' \le s$, so $G_s$ does not depend on $h_{s'}$ for $s' > s$. The characteristic polynomial of a block-triangular matrix is the product of those of its diagonal blocks.
\end{proof}

On the materialized Jacobian the strictly-upper token blocks vanish identically, and the spectrum equals the union of the diagonal blocks' spectra to within $4 \times 10^{-8}$. At a fixed operating point, attention's cross-position influence is therefore spectrally invisible: the strictly-lower blocks contribute only to the operator norm and to transients, and attention shapes the spectrum only indirectly, by moving $h^*$ and hence the diagonal blocks. The cross-token mass is small, $\approx 20\%$ of the increment's operator norm. Over a finite loop depth $T$, sensitivity is governed by the product $J_G(h_{T-1}) \cdots J_G(h_0)$ along the trajectory, which non-normal transients can inflate, rather than by the spectral margin.

\paragraph{Connection to Parcae's test-time scaling law.} Parcae~\citep[\S5.3]{parcae2026} fits $\mathcal{L}(T) = \mathcal{L}_\infty + Ze^{-zT}$. The fixed-point iteration converges at an asymptotic rate $\rho_{\text{spec}}(J_G)$, which exceeds the carry $\rho$ at every operating point and rises toward $1$ as $\rho \to 1$, so each iteration refines the state by a factor $\approx \rho_{\text{spec}}$. That the loss inherits a single geometric mode from the state is a hypothesis: a loss expansion can mix several decay modes and transients, and averaging over inputs adds more.

\paragraph{What the diagonal constraint does and does not guarantee.} By \cref{prop:existence}, $\rho(\bar{A}) < 1$ guarantees bounded trajectories and the existence of a fixed point. It does not guarantee convergence to one: some initializations satisfy it by construction yet do not settle within $2 \times 10^4$ iterations, because the dense nonlinear term sustains a bounded orbit. Nor is it necessary for local stability, since $J_G = D + J_f$ can have all its eigenvalues inside the unit circle even with $\rho(D) > 1$ in some channel; without the constraint, what is lost is the global boundedness guarantee. The useful training-time diagnostic is therefore the loop's own convergence, the residual $\|G(h) - h\|$ across iterations, together with $\rho_{\text{spec}}(J_G)$ at the settled state, rather than $\rho(\bar{A}) < 1$ alone.

\paragraph{Width.} Are the high-carry margin collapse and the non-settling minority small-width artifacts? We repeat the random-weight protocol at $d \in \{32, 64, 128, 256\}$ (head dimension $8$, seq $16$, $100$ seeds per width, $\rho = 0.9$). At every settled state we materialize $J_G$ and take $\rho_{\text{spec}}$ exactly from the per-token diagonal blocks (\cref{prop:block-triangular}) and $\|J_G\|_2$ by SVD (\cref{fig:width}; per-width numbers in \cref{tab:width}, \cref{app:width}). Over the converged seeds, every exact margin is positive at every width ($0$ of $285$ negative) and every operator norm exceeds $1$, so the picture of \cref{tab:seed-robustness} holds per instance up to $d = 256$. The mean margin falls by about a third per doubling of width, roughly $d^{-2/3}$; settle times grow in step, the critical slowing down a vanishing margin predicts; and the non-settling fraction rises from $17$--$20\%$ to $29\%$ at $d = 256$. The collapse is specific to high carry: on a ten-seed companion ladder the $\rho = 0.3$ margin stays $\approx 0.10$ through $d = 256$, with $\|J_G\|_2 \approx 1.8$--$2.3$.

\begin{figure}[ht]
\centering
\begin{minipage}{0.48\textwidth}
\centering
\begin{tikzpicture}
\begin{loglogaxis}[
    width=\textwidth, height=0.8\textwidth,
    xlabel={width $d$}, ylabel={mean spectral margin},
    xmin=24, xmax=340, ymin=0.002, ymax=0.03,
    xtick={32,64,128,256}, xticklabels={32,64,128,256},
    ytick={0.002,0.005,0.01,0.02}, yticklabels={0.002,0.005,0.01,0.02},
    grid=major, grid style={gray!25},
    tick label style={font=\footnotesize}, label style={font=\footnotesize},
    legend pos=south west, legend style={font=\scriptsize},
]
\addplot[gray, dashed, domain=24:340, samples=2] {0.0206*(32/x)^(2/3)};
\addlegendentry{$\propto d^{-2/3}$}
\addplot[blue, thick, mark=*, mark size=1.8pt, error bars/.cd, y dir=both, y explicit]
coordinates {(32,0.0206) +- (0,0.0012) (64,0.0145) +- (0,0.0009) (128,0.0089) +- (0,0.0006) (256,0.0056) +- (0,0.0005)};
\addlegendentry{exact, converged seeds ($\pm$ s.e.)}
\addplot[red!80!black, thick, mark=triangle*, mark size=2pt, densely dashed] coordinates {(32,0.0177) (64,0.0124) (128,0.0063) (256,0.0044)};
\addlegendentry{implied by settle time}
\end{loglogaxis}
\end{tikzpicture}
\end{minipage}\hfill
\begin{minipage}{0.48\textwidth}
\centering
\begin{tikzpicture}
\begin{semilogxaxis}[
    width=\textwidth, height=0.8\textwidth,
    xlabel={width $d$}, ylabel={non-settling fraction},
    xmin=24, xmax=340, ymin=0, ymax=0.5,
    xtick={32,64,128,256}, xticklabels={32,64,128,256},
    grid=major, grid style={gray!25},
    tick label style={font=\footnotesize}, label style={font=\footnotesize},
]
\addplot[red!80!black, thick, mark=square*, mark size=1.8pt, error bars/.cd, y dir=both, y explicit]
coordinates {(32,0.17) +- (0.085,0.061) (64,0.20) +- (0.089,0.067) (128,0.19) +- (0.088,0.065) (256,0.29) +- (0.095,0.080)};
\end{semilogxaxis}
\end{tikzpicture}
\end{minipage}
\caption{Width ladder at $\rho = 0.9$ (\cref{tab:width}). Left: mean exact spectral margin over converged seeds, falling by about a third per doubling of width, with the margin implied by the mean settle time. Right: fraction of seeds that do not settle within $2 \times 10^4$ iterations, with Wilson $95\%$ intervals.}
\label{fig:width}
\end{figure}

\subsection{A Slow Channel Narrows the Margin}
\label{sec:per-channel}

With a learnable vector $\lambda$, each channel has its own carry $\bar{A}_i$ and memory horizon $\tau_i = 1/(1-\bar{A}_i)$, the geometric-sum gain (the e-folding time $-1/\ln \bar{A}_i$ agrees with it as $\bar{A}_i \to 1$). Fast channels ($\bar{A}_i \approx 0.3$, $\tau \approx 1.4$) track the latest refinement; slow channels ($\bar{A}_i \approx 0.99$, $\tau \approx 100$) integrate long-range state. The stability margin, however, is governed by $\max_i \bar{A}_i$ in every profile we measure (\cref{tab:per-channel}): a single channel at $0.99$ collapses the margin by an order of magnitude regardless of the other $31$, and the settle count rises accordingly. The operator norm is complementary: at uniform $0.5$ the dense increment dominates and $\|J_G\|_2 = 1.48$, while in every profile with a $0.99$ channel the near-normal carry dominates and $\|J_G\|_2 \approx \rho_{\text{spec}} \approx 1$. In two of these profiles the operator norm is slightly below $1$ ($0.992$ and $0.997$), so the scalar-carry finding that it always exceeds $1$ does not extend to them.

\begin{table}[ht]
\centering
\small
\begin{tabular}{@{}lcccccc@{}}
\toprule
Carry profile & Mean $\bar{A}$ & Max $\bar{A}$ & $\rho_{\text{spec}}(J_G)$ & $\|J_G\|_2$ & margin & settle iters \\
\midrule
Uniform 0.5 (scalar) & 0.50 & 0.50 & 0.901 & 1.481 & 0.099 & 168 \\
31 channels at 0.5, 1 at 0.99 & 0.515 & 0.99 & 0.991 & 0.992 & 0.009 & 1535 \\
Ramp 0.5 $\to$ 0.99 & 0.745 & 0.99 & 0.989 & 0.997 & 0.011 & 1326 \\
Uniform 0.99 (scalar) & 0.99 & 0.99 & 0.997 & 1.003 & 0.003 & 5014 \\
\bottomrule
\end{tabular}
\caption{Per-channel carry profiles (same block and protocol as \cref{tab:cf}, direct eigendecomposition and SVD at verified fixed points). In these four profiles, the slowest channel pins the spectral margin.}
\label{tab:per-channel}
\end{table}

The channel with the longest memory horizon is the channel with the tightest stability margin. This is the design tension of diagonal state-space models (S4, S4D, Mamba), in which the eigenvalue closest to the unit circle provides the most memory and the least margin, transplanted from the sequence axis to the depth axis. The training diagnostic that follows is to monitor $\max_i \bar{A}_i$, not the mean.

\section{Training Dynamics}
\label{sec:training}

The results so far are established with random weights. This section asks what gradient descent does with the two paths. Without the carry constraint, gradient descent does drive the carry past $1$ at high learning rates, and the resulting state explosion is hidden from the loss by the readout's final normalization; this reproduces Parcae's instability at toy scale (\cref{subsec:unconstrained}). With the constraint in place, we test whether training recruits the carry on a task that forces memory across iterations (\cref{subsec:windowed}) and isolate the depth-memory mechanism by ablation (\cref{subsec:ablation}). We then test by intervention whether any carry value is necessary at fixed depth (\cref{subsec:intervention}) and under sampled depth (\cref{subsec:sampled-depth}), and close with a constructed task on which the carry is recruited (\cref{subsec:boundary}). Throughout, \emph{recruited} means that training moves $\bar{A}$ substantially from its initialization in the direction the task's memory demand predicts.

\subsection{Windowed Attention: The Carry Is Not Recruited}
\label{subsec:windowed}

Two preliminary tasks put little or no demand on depth memory (\cref{app:training}). A character-level model on Tiny Shakespeare barely moves the carry ($\max \bar{A}$ from $0.19$ to $0.23$ in one short run), and a copy task under full causal attention is bridged by attention in a single hop, whatever the gap. The carry persists state across loop iterations at a fixed position and cannot move information across positions, so forcing depth memory requires restricting attention.

We restrict attention to window $w = 1$, so each position attends only to itself and its predecessor, and moving information across a gap of $k$ positions takes about $k$ loop iterations. The sequence is $[A][\text{pad} \times k][?]$: the symbol $A$, one of 14 content tokens (vocabulary 16), sits at position $0$ and must be reproduced at the query position $k+1$; chance accuracy is $\approx 0.07$. \Cref{fig:relay} shows the mechanics at $k = 2$. In the first iteration position $1$ reads $A$ from position $0$, in the second position $2$ reads it from position $1$, and so on, so the symbol reaches the query after $k+1$ iterations. Each hop reads the previous position's state from the preceding iteration, so the state must carry the symbol from one iteration to the next. Because the source is re-injected every iteration, a fresh wave starts every iteration, and nothing has to be held for longer than one. The question is which path carries the state across an iteration.

If a single channel of the carry held the symbol for $k$ iterations, one would expect $\max(\bar{A}) \approx 1 - 1/k$. This is the strongest carry-based hypothesis, not the only one, since a relay need only hold its packet for the iteration or two between write and read. We report it for orientation; the intervention of \cref{subsec:intervention} does not depend on it.

\begin{figure}[ht]
\centering
\begin{tikzpicture}[>=stealth, font=\small]
\node[font=\small\bfseries] at (0,0.75) {$[A]$};
\node[font=\small\bfseries] at (1.6,0.75) {$[p]$};
\node[font=\small\bfseries] at (3.2,0.75) {$[p]$};
\node[font=\small\bfseries] at (4.8,0.75) {$[?]$};
\node[anchor=east, gray, font=\footnotesize] at (-0.8,0) {$t=0$ (input)};
\node[anchor=east, gray, font=\footnotesize] at (-0.8,-1.0) {after loop $1$};
\node[anchor=east, gray, font=\footnotesize] at (-0.8,-2.0) {after loop $2$};
\node[anchor=east, gray, font=\footnotesize] at (-0.8,-3.2) {after loop $3$};
\node[draw, circle, fill=blue!20, minimum size=0.55cm, inner sep=0pt] (a0) at (0,0.0) {$A$};
\node[draw, circle, fill=blue!20, minimum size=0.55cm, inner sep=0pt] (a1) at (1.6,-1.0) {$A$};
\node[draw, circle, fill=blue!20, minimum size=0.55cm, inner sep=0pt] (a2) at (3.2,-2.0) {$A$};
\node[draw, circle, fill=blue!20, minimum size=0.55cm, inner sep=0pt] (a3) at (4.8,-3.2) {$A$};
\draw[->, thick, blue!60!black] (a0) -- (a1);
\node[font=\footnotesize, text=blue!60!black, align=left, anchor=north west] at (-0.3,-2.75) {attention hops (solid):\\$w = 1$, one position\\per loop iteration};
\draw[->, thick, blue!60!black] (a1) -- (a2);
\draw[->, thick, blue!60!black] (a2) -- (a3);
\draw[->, thick, dashed, red!70!black] (a3) -- (4.8,-4.3) node[right=2pt, pos=0.5, font=\footnotesize, text=red!70!black, align=left] {carried across each iteration\\by which path?};
\node[font=\footnotesize, gray] at (4.8,-4.6) {readout at $T = 8$};
\end{tikzpicture}
\caption{Windowed copy at $k=2$; each row is the state after the labeled loop iteration. Each attention hop (solid) moves the symbol one position per iteration, so it reaches the query after loop $3$, and a fresh wave starts every iteration because the source is re-injected. Between iterations the symbol lives in a position's hidden state (dashed); which path carries it across is the question.}
\label{fig:relay}
\end{figure}

\Cref{app:training} lists all hyperparameters: $d = 32$, $4$ heads, sequence length $12$, loop depth $T = 8$, Adam, the carry initialized at $\bar{A} = 0.5$, and the curriculum $k = 2, \dots, 6$ with each stage warm-started from the previous one; accuracy is the argmax at the query over $64$ fresh draws.

In a single run through the curriculum, every stage is solved (accuracy $0.91$ at $k = 2$ and $1.00$ at $k = 3$--$6$), so the loop can relay information across iterations through windowed attention. The carry is not recruited: $\max(\bar{A})$ drifts from $0.524$ down to $0.486$ as $k$ grows, against the heuristic prediction of $0.833$ (\cref{tab:windowed-copy}). Once a stage is solved, the mean gradient $\partial L / \partial \bar{A}$ is positive but of order $10^{-5}$, so the finding rests on the level of $\bar{A}$, not on the gradient's sign. The curriculum had to step $k$ by one; a jump from $k = 4$ to $6$ did not train. The non-recruitment replicates across five training seeds (\cref{subsec:intervention}).

\subsection{Ablation: The Nonlinear Path Carries the Relay}
\label{subsec:ablation}

To attribute the depth memory, we cut the block's access to $h_t$: the block sees only $e$, which leaves the carry as the only path across iterations (\cref{fig:ablation} in \cref{app:training}). The ablated recurrence is $h_{t+1} = D h_t + f(e)$, so $h_T = D^T h_0 + \sum_{j<T} D^j f(e)$. Because $D$ is diagonal, iterating cannot widen the receptive field of a single block application, so under windowed attention the ablated model cannot reach the source token at all. We compare three conditions at $k = 4$ ($T = 8$, $d = 24$, cold start). The full windowed model solves the task ($1.00$). The ablated windowed model stays at chance ($0.09$), as the receptive-field argument requires, and its carry does not rise to compensate ($\max(\bar{A}) = 0.52$), nor could it. The ablated model with full attention solves the task ($1.00$), so the ablation removed multi-iteration propagation, not capacity. The nonlinear path is therefore \emph{necessary} for multi-hop propagation; whether the carry's resting value contributes to the full model's solution is what the next intervention tests.

\subsection{Intervention: Freezing and Overriding the Carry}
\label{subsec:intervention}

Does the solution the full model finds depend on the carry's resting value? We intervene on the carry directly in the configuration of \cref{subsec:windowed} ($T = 8$, $w = 1$, $d = 32$, the same curriculum, identical model initialization and training batch stream across conditions). Four training conditions: (a) learned carry; (b) frozen at its initialization $\bar{A} = 0.5$; (c) frozen at $0.02$; (d) frozen at exactly $0$, i.e.\ $h_{t+1} = f(h_t, e)$ with no linear carry at all, since in a normalized system a numerically small pathway is not automatically negligible.

\begin{table}[ht]
\centering
\begin{tabular}{@{}lcccccc@{}}
\toprule
Condition & $k=2$ & $k=3$ & $k=4$ & $k=5$ & $k=6$ & final $\max(\bar{A})$ \\
\midrule
(a) learned carry & 0.94 & 1.00 & 1.00 & 1.00 & 1.00 & 0.487 \\
(b) frozen $\bar{A} = 0.5$ & 1.00 & 1.00 & 1.00 & 1.00 & 1.00 & 0.500 \\
(c) frozen $\bar{A} = 0.02$ & 1.00 & 1.00 & 1.00 & 1.00 & 1.00 & 0.020 \\
(d) frozen $\bar{A} = 0$ & 1.00 & 1.00 & 1.00 & 1.00 & 1.00 & 0.000 \\
\bottomrule
\end{tabular}
\caption{Carry-freezing intervention on windowed copy. Each entry is the accuracy at the end of that curriculum stage (64 fresh draws), not the final model's accuracy on that gap (\cref{tab:retention}). All three frozen conditions solve every stage. Condition (a) repeats the \cref{tab:windowed-copy} configuration from the same initialization; the evaluation draws differ, and the batch streams diverge after the first stage (the earlier run draws one extra batch per stage for its gradient measurement), hence $0.94$ vs.\ $0.91$ at $k=2$ and $0.487$ vs.\ $0.486$ for the final $\max(\bar{A})$.}
\label{tab:intervention}
\end{table}

All three frozen conditions solve every stage (\cref{tab:intervention}), and \cref{fig:carry-trace} shows the learned carry staying within $0.03$ of its initialization throughout while the loss falls to the same level in all four conditions. At fixed depth, then, the carry is dispensable for learning this task: the multi-hop memory needs no linear retention at all, and the relay of \cref{fig:relay} runs through the block.

The trained model nonetheless uses the carry at run time. Taking the trained model of condition (a) and setting $\bar{A} := 0$ without retraining collapses accuracy to chance at every gap ($1.00 \to 0.14$ at $k = 6$); restoring $\bar{A} := 0.5$ recovers it (\cref{tab:retention}). The trained weights are therefore calibrated to the carry value seen in training, but training neither tunes that value nor needs any particular one to learn the task from scratch. The override shows dependence, not its nature: zeroing the carry removes both any retention it implements and a scale the block's weights are calibrated against. What rules out a \emph{necessary} memory role is the zero-carry retraining.

\begin{figure}[ht]
\centering
\begin{minipage}{0.48\textwidth}
\centering
\begin{tikzpicture}
\begin{axis}[
    width=0.94\textwidth, height=0.75\textwidth,
    xlabel={training step (stages $k = 2$ to $6$)}, ylabel={$\max(\bar{A})$},
    xmin=0, xmax=1400, ymin=-0.05, ymax=0.9,
    grid=major, grid style={gray!25},
    tick label style={font=\footnotesize}, label style={font=\footnotesize},
    legend style={at={(0.98,0.74)}, anchor=north east, font=\scriptsize, fill opacity=0.9, text opacity=1},
]
\foreach \x in {300,500,700,950} {\addplot[gray, dotted, forget plot] coordinates {(\x,-0.05) (\x,0.9)};}
\addplot[gray!60, dashed, forget plot] coordinates {(0,0.833) (1400,0.833)};
\node[font=\scriptsize, gray!60!black, anchor=south west] at (axis cs:20,0.84) {heuristic $1 - 1/6$};
\addplot[blue, thick, mark=*, mark size=1.2pt] coordinates {(0,0.500) (100,0.512) (200,0.520) (300,0.524) (400,0.520) (500,0.525) (600,0.525) (700,0.527) (800,0.517) (900,0.518) (1050,0.487) (1150,0.487) (1250,0.487) (1350,0.487)};
\addlegendentry{(a) learned}
\addplot[red!80!black, thick] coordinates {(0,0.5) (1350,0.5)};
\addlegendentry{(b) frozen $0.5$}
\addplot[orange!80!black, thick] coordinates {(0,0.02) (1350,0.02)};
\addlegendentry{(c) frozen $0.02$}
\addplot[green!50!black, thick] coordinates {(0,0.0) (1350,0.0)};
\addlegendentry{(d) frozen $0$}
\end{axis}
\end{tikzpicture}
\end{minipage}\hfill
\begin{minipage}{0.48\textwidth}
\centering
\begin{tikzpicture}
\begin{semilogyaxis}[
    width=0.94\textwidth, height=0.75\textwidth,
    xlabel={training step}, ylabel={training loss},
    xmin=0, xmax=1400, ymin=0.002, ymax=4,
    grid=major, grid style={gray!25},
    tick label style={font=\footnotesize}, label style={font=\footnotesize},
    legend pos=north east, legend style={font=\scriptsize, fill opacity=0.9, text opacity=1},
]
\foreach \x in {300,500,700,950} {\addplot[gray, dotted, forget plot] coordinates {(\x,0.002) (\x,4)};}
\addplot[blue, thick, mark=*, mark size=1.2pt] coordinates {(100,1.608) (200,0.857) (300,0.349) (400,0.125) (500,0.049) (600,0.053) (700,0.029) (800,0.043) (900,0.021) (1050,0.040) (1150,0.025) (1250,0.014) (1350,0.013)};
\addlegendentry{(a) learned}
\addplot[red!80!black, thick, mark=square*, mark size=1.1pt] coordinates {(100,1.686) (200,0.950) (300,0.251) (400,0.132) (500,0.049) (600,0.040) (700,0.023) (800,0.075) (900,0.026) (1050,0.065) (1150,0.023) (1250,0.017) (1350,0.012)};
\addlegendentry{(b) frozen $0.5$}
\addplot[orange!80!black, thick, mark=triangle*, mark size=1.4pt] coordinates {(100,1.333) (200,0.422) (300,0.106) (400,0.047) (500,0.024) (600,0.021) (700,0.012) (800,0.010) (900,0.007) (1050,0.172) (1150,0.037) (1250,0.012) (1350,0.012)};
\addlegendentry{(c) frozen $0.02$}
\addplot[green!50!black, thick, mark=diamond*, mark size=1.4pt] coordinates {(100,1.440) (200,0.335) (300,0.108) (400,0.044) (500,0.024) (600,0.019) (700,0.015) (800,0.013) (900,0.009) (1050,0.016) (1150,0.007) (1250,0.005) (1350,0.004)};
\addlegendentry{(d) frozen $0$}
\end{semilogyaxis}
\end{tikzpicture}
\end{minipage}
\caption{Training traces of the four intervention conditions (\cref{tab:intervention}), logged every $100$ steps; dotted lines mark the stage boundaries. Left: $\max(\bar{A})$. The learned carry stays within $0.03$ of its initialization, far below the single-channel-retention heuristic $1 - 1/k$ (dashed, at $k = 6$). Right: training loss, which reaches the same level in all four conditions.}
\label{fig:carry-trace}
\end{figure}

The final models also differ in how much of the earlier curriculum gaps they retain, with the zero-carry models retaining the most (\cref{app:retention}).

\paragraph{Seed replication.} We replicated conditions (a) and (d) across training seeds, each seed changing both the initialization and the batch stream. The learned condition replicates in $5$ of $5$ seeds: every stage solves ($k \ge 3$ at $1.00$), final $\max(\bar{A})$ spans $0.487$--$0.512$, and the mean $\bar{A}$ declines from $0.50$ to $0.43$--$0.46$ in all five. The zero-carry condition replicates in $9$ of $10$ seeds, each solving every stage. The tenth fails every stage: its loss stays on a plateau for the whole run, and for three times as long on $k = 2$ alone, while the learned-carry condition at the same initialization solves everything. Any single change of batch order, initialization, or learning rate ($3 \times 10^{-4}$ to $3 \times 10^{-3}$) rescues it. The failure is thus an optimization effect of one exact configuration, for which the carry is one of several interchangeable remedies. This is consistent with the carry easing optimization the way a skip connection~\citep{he2016} does, here along the depth axis.

\paragraph{Stability after training.} On two fresh $k = 6$ inputs we settle the loop under the trained weights of condition (a) and materialize the dense Jacobian. At the settled state the trained model's spectral radius is \emph{smaller} than at initialization, $\rho_{\text{spec}} = 0.722$/$0.728$ vs.\ $0.842$/$0.843$, and settling from the $T = 8$ state takes $56$--$64$ iterations vs.\ $104$. At the $T = 8$ operating state itself, which is not settled, the local radius exceeds $1$ ($1.24$/$1.66$; operator norm $2.98$/$2.62$): the trajectory passes through a transiently expansive region before contracting.

\paragraph{Depth.} Evaluating the trained model at other loop depths shows that the computation lives in a narrow window around the training depth (\cref{fig:depth-sweep}, left). Accuracy on $k = 6$ is $1.00$ at $T = 8$, $0.89$ at $T = 9$, and $0.42$ at $T = 10$; it is at chance below $T = 7$ and from $T = 12$ on, and stays there once the loop has settled. The answer is thus read out from a transient, not from an equilibrium: the fixed point the model settles into is more stable than at initialization, yet the trained readout cannot recover the answer from it (a separately trained probe might). This is expected of a model trained at a single depth, without the depth sampling that Parcae and \citet{geiping2025} use to make the readout depth-robust. The degradation under excess iterations is the overthinking of \citet{bansal2022}; \citet{kohli2026} report it in recurrent-depth transformers and find dynamic-depth training more robust to it. Fixed-point stability and finite-depth performance are thus distinct: training improved the former while placing the computation in the transient.

\begin{figure}[ht]
\centering
\begin{minipage}{0.48\textwidth}
\centering
\begin{tikzpicture}
\begin{axis}[
    width=0.95\textwidth, height=0.68\textwidth,
    title={\footnotesize fixed-depth training ($T = 8$)},
    xlabel={loop depth $T$ at evaluation},
    ylabel={accuracy},
    xmin=0.5, xmax=16.5, ymin=0, ymax=1.05,
    xtick={2,4,...,16},
    legend style={at={(0.5,-0.3)}, anchor=north, font=\scriptsize, legend columns=2},
    grid=major, grid style={gray!25},
    tick label style={font=\footnotesize},
    label style={font=\footnotesize},
]
\addplot[gray, dashed, sharp plot, mark=none] coordinates {(0.5,0.071) (16.5,0.071)};
\addlegendentry{chance}
\addplot[blue, thick, mark=*, mark size=1.8pt] coordinates {
    (1,0.0625) (2,0.0625) (3,0.0625) (4,0.0625) (5,0.0625) (6,0.0625) (7,0.359) (8,1.0) (9,0.891) (10,0.422) (11,0.141) (12,0.078) (14,0.078) (16,0.078)
};
\addlegendentry{$k = 6$, learned carry}
\addplot[red!80!black, thick, mark=square*, mark size=1.6pt] coordinates {
    (1,0.047) (2,0.047) (3,0.047) (4,0.047) (5,0.047) (6,0.3125) (7,0.344) (8,0.484) (9,0.516) (10,0.3125) (11,0.156) (12,0.078) (14,0.078) (16,0.078)
};
\addlegendentry{$k = 5$, learned carry}
\addplot[black, dotted, thick, sharp plot, mark=none] coordinates {(8,0) (8,1.05)};
\addlegendentry{training depth}
\end{axis}
\end{tikzpicture}
\end{minipage}\hfill
\begin{minipage}{0.48\textwidth}
\centering
\begin{tikzpicture}
\begin{axis}[
    width=0.95\textwidth, height=0.68\textwidth,
    title={\footnotesize sampled-depth training ($T \sim \mathcal{U}\{7, \dots, 16\}$)},
    xlabel={loop depth $T$ at evaluation},
    ylabel={accuracy},
    xmin=0.5, xmax=24.5, ymin=0, ymax=1.05,
    xtick={2,4,...,24},
    legend style={at={(0.5,-0.3)}, anchor=north, font=\scriptsize, legend columns=2},
    grid=major, grid style={gray!25},
    tick label style={font=\footnotesize},
    label style={font=\footnotesize},
]
\addplot[gray!20, fill=gray!20, draw=none, forget plot] coordinates {(6.5,0) (16.5,0) (16.5,1.05) (6.5,1.05)} \closedcycle;
\addplot[gray, dashed, sharp plot, mark=none] coordinates {(0.5,0.071) (24.5,0.071)};
\addlegendentry{chance}
\addplot[blue, thick, mark=*, mark size=1.8pt] coordinates {
    (4,0.109) (6,0.109) (7,0.828) (8,0.9375) (9,1.0) (10,1.0) (12,0.9375) (16,0.703) (24,0.0)
};
\addlegendentry{$k = 6$, learned carry}
\addplot[red!80!black, thick, mark=square*, mark size=1.6pt] coordinates {
    (4,0.0625) (6,0.109) (7,0.609) (8,0.625) (9,0.781) (10,0.9375) (12,1.0) (16,1.0) (24,0.234)
};
\addlegendentry{$k = 6$, frozen carry $0.5$}
\addlegendimage{area legend, draw=none, fill=gray!20}
\addlegendentry{sampled training depths}
\end{axis}
\end{tikzpicture}
\end{minipage}
\caption{Depth sweeps of the final models after the $k = 6$ stage (64 fresh draws per point; dashed line: chance). Left: fixed-depth training, condition (a); accuracy on $k = 6$ peaks at the training depth and is at chance two iterations below it and from four iterations above it. Right: sampled-depth training (\cref{tab:stochastic-depth}); accuracy is high across the sampled range (shaded) and degrades at the depths evaluated outside it.}
\label{fig:depth-sweep}
\end{figure}

\subsection{Sampled Depth: The Carry's Magnitude Matters}
\label{subsec:sampled-depth}

Parcae and \citet{geiping2025} train with the loop depth sampled (Parcae per sequence within each micro-batch, from a Poisson-type distribution), a protocol meant to mitigate the depth window of \cref{fig:depth-sweep}. We repeated the intervention under a simplified version: one depth per batch, drawn uniformly from $\{7, \dots, 16\}$ at every step, with the same depth sequence in every condition and everything else as in \cref{tab:intervention}; accuracy is reported at $T = 8$ after each stage (\cref{tab:stochastic-depth}). The learned carry trains, more slowly than at fixed depth ($0.94$ at $k = 6$ on seed $0$, $0.66$ on a second seed), and its final model reads out across the sampled range and no further (\cref{fig:depth-sweep}, right). In both seeds the carry moves as little as at fixed depth ($\max(\bar{A}) \le 0.56$). The carry frozen at $0.5$ also trains, though unevenly across stages and depths. The two small carries do not: the zero carry stays at chance after every stage on four seeds and under the learning-rate change that repaired its fixed-depth failure, and the carry frozen at $0.02$, which solved every stage at fixed depth, also stays at chance.

Dispensability is therefore a fixed-depth result. Under sampled depth the block must implement the relay at every depth it is trained at, and in this configuration gradient descent does not find such a solution without a substantial carry. Six failing runs do not decide whether that is an optimization effect, as the fixed-depth failure was, or a representational one. That a frozen carry suffices, and that the learned carry is still not recruited, favors the former: the carry would ease optimization as a skip connection along depth does, and a $0.02$ skip is too weak to help.

\begin{table}[ht]
\centering
\begin{tabular}{@{}lcccccc@{}}
\toprule
Condition & $k=2$ & $k=3$ & $k=4$ & $k=5$ & $k=6$ & final $\max(\bar{A})$ \\
\midrule
(a) learned carry, seed 0 & 0.56 & 0.73 & 1.00 & 0.95 & 0.94 & 0.559 \\
(a) learned carry, seed 1 & 0.09 & 0.25 & 0.42 & 0.72 & 0.66 & 0.534 \\
(b) frozen $\bar{A} = 0.5$ & 0.52 & 0.48 & 0.17 & 0.95 & 0.63 & 0.500 \\
(c) frozen $\bar{A} = 0.02$ & 0.09 & 0.05 & 0.02 & 0.08 & 0.11 & 0.020 \\
(d) frozen $\bar{A} = 0$, seed 0 & 0.09 & 0.06 & 0.02 & 0.08 & 0.11 & 0.000 \\
(d) frozen $\bar{A} = 0$, seed 1 & 0.06 & 0.03 & 0.06 & 0.03 & 0.06 & 0.000 \\
(d) frozen $\bar{A} = 0$, seed 2 & 0.05 & 0.06 & 0.06 & 0.03 & 0.08 & 0.000 \\
(d) frozen $\bar{A} = 0$, seed 3 & 0.11 & 0.11 & 0.13 & 0.16 & 0.06 & 0.000 \\
(d) seed 0, learning rate $3 \times 10^{-4}$ & 0.11 & 0.16 & 0.14 & 0.22 & 0.20 & 0.000 \\
\bottomrule
\end{tabular}
\caption{Carry intervention under sampled depth: the \cref{tab:intervention} protocol with the loop depth drawn per batch from $\{7, \dots, 16\}$. Each entry is the accuracy at $T = 8$ at the end of that curriculum stage (64 fresh draws; chance $\approx 0.07$). The seed changes the initialization and the batch stream; the depth sequence is the same in every run. Carries of $0.5$ train; $0.02$ and $0$ do not.}
\label{tab:stochastic-depth}
\end{table}

\subsection{Boundary: Carry Recruited for a Channelwise Gain}
\label{subsec:boundary}

Does a task exist on which the carry's diagonal structure is the efficient path? We constructed one: a single position ($n = 1$) with channel values $u_c \sim \mathcal{N}(0,1)$ held constant across $T = 8$ iterations, and a regression target $Y_c = u_c \sum_{t=0}^{T-1} (a^*_c)^t$ per channel, exactly the state a carry with $\bar{A}_c = a^*_c$ and a pass-through block reaches after $T$ iterations ($Y_c \approx 1.43\,u_c$ at $a^*_c = 0.3$, $5.70\,u_c$ at $0.9$). Target rates ramp across the $d = 16$ channels, $a^*_c \in [0.30, 0.90]$; $\bar{A}$ is initialized at $0.5$ and the loss is an MSE over channels (one run).

Here $\bar{A}$ moves substantially and co-varies with the target: $\text{corr}(\bar{A}_c, a^*_c) = 0.92$ over the 16 channels, with fit slope $0.81$ (\cref{fig:diagonal-memory}). The agreement is co-variation, not numerical: the high-target channels stop at $\bar{A} = 0.74$ against $0.90$, and the $a^*_c = 0.50$ channel ends at $0.22$, below both its target and its initialization and below the $0.30$-target channel, so the block supplies part of the required gain and the per-channel allocation is noisy. Because the input is constant across iterations, the target at fixed $T$ is equivalent to a per-channel \emph{gain}; the experiment identifies which path implements that gain (the carry), not that the learned $\bar{A}_c$ traces a decay trajectory.

The task is artificial by design (regression head, sequence length 1, direct injection, an MSE loss giving each $\bar{A}_c$ a direct gradient) and maps the boundary of the claim. Because it changes several things at once relative to the copy tasks, it is a contrasting case in which the carry is recruited, not a controlled isolation of the recruiting factor.

\begin{figure}[ht]
\centering
\begin{tikzpicture}
\begin{axis}[
    width=0.65\textwidth,
    height=0.5\textwidth,
    xlabel={Target $a^*_c$},
    ylabel={Learned $\bar{A}_c$},
    xmin=0.25, xmax=0.95,
    ymin=0.15, ymax=0.85,
    grid=major,
    grid style={gray!30},
    legend pos=north west,
]
\addplot[black, dashed, thin, domain=0.25:0.95] {x};
\addlegendentry{Identity}
\addplot[red, thick, domain=0.25:0.95] {0.805*x - 0.011};
\addlegendentry{Least-squares fit ($r = 0.92$)}
\addplot[only marks, mark=o, mark size=2.5pt, blue, thick] coordinates {
    (0.30, 0.342) (0.34, 0.305) (0.38, 0.283) (0.42, 0.270)
    (0.46, 0.407) (0.50, 0.222) (0.54, 0.357) (0.58, 0.532)
    (0.62, 0.458) (0.66, 0.499) (0.70, 0.514) (0.74, 0.616)
    (0.78, 0.645) (0.82, 0.679) (0.86, 0.687) (0.90, 0.737)
};
\addlegendentry{Learned $\bar{A}_c$}
\end{axis}
\end{tikzpicture}
\caption{Channelwise gain task: learned carry $\bar{A}_c$ tracks target decay $a^*_c$ with correlation 0.92 over the 16 channels. This is co-variation, not numerical agreement: the fit slope is $0.81$ and high-target channels stop short of their targets (see text). This is the boundary case where the carry's diagonal structure matches the task structure.}
\label{fig:diagonal-memory}
\end{figure}

\subsection{Synthesis}

\Cref{tab:synthesis} summarizes the training experiments. On the cross-channel tasks, $\max(\bar{A})$ stays near its initialization whatever the memory demand, and the intervention sharpens this non-recruitment into dispensability at fixed depth. Under sampled depth, carries of $0$ and $0.02$ fail to train while $0.5$, frozen or learned, suffices, and the learned carry is still not recruited. Only on the per-channel task is the carry recruited.

\begin{table}[ht]
\centering
\small
\begin{tabular}{@{}llll@{}}
\toprule
Experiment & Final $\max(\bar{A})$ & Carry moved by training? & Task learned? \\
\midrule
Tiny Shakespeare & 0.23 & little & --- (language model) \\
Windowed copy ($k{=}6$), learned carry & 0.49 & no & yes \\
Ablation (block blind to $h_t$) & 0.52 & no & no (unreachable) \\
Windowed copy, carry frozen & 0.5 / 0.02 / 0 & frozen & yes, all three \\
Sampled depth, learned / frozen & 0.56 / 0.5 / 0.02 / 0 & no / frozen & 0.5: yes; 0.02, 0: no \\
Channelwise gain & 0.22--0.74 & yes ($r{=}0.92$) & yes \\
\bottomrule
\end{tabular}
\caption{Carry behavior across the training experiments. ``Moved'' means a substantial change from the initialization $0.5$ (or $0.19$ for Tiny Shakespeare) in the direction the task's memory demand predicts.}
\label{tab:synthesis}
\end{table}

The looped transformer has two depth-memory paths: the linear carry (diagonal, stability-constrained) and the nonlinear recurrence through $f(h_t, e)$ (dense, high-capacity). Comparing their expressivity only makes sense on the bounded set of states the dynamics visit. The increment is uniformly bounded (\cref{prop:existence}), so the nonlinear path cannot represent the carry's own unbounded map $h \mapsto \bar{A} \odot h$ globally. On the bounded set the dynamics visit, though, the block's dense nonlinear class is far richer than per-channel rescaling, and the carry cannot express cross-channel interactions at any scale. The increment also has an exact blind spot: by \cref{prop:mean-shift} it ignores per-token constant shifts, to which the carry responds. With scalar carry that direction decays at exactly rate $\rho$ and the readout discards it; with non-uniform carry it feeds back into the centered dynamics. Finally, the block-based solution carried no measured stability cost: on the trained windowed-copy model the settled-state spectral radius falls relative to initialization ($0.72$ vs.\ $0.84$, two inputs of one model).

Parcae's carry constraint restricts a degree of freedom that, on these tasks at fixed depth, the optimizer does not need to learn the memory task. The constraint prevents the instability of \cref{tab:unconstrained}, and on cross-channel workloads at fixed depth it appears to cost nothing. Under sampled depth the zero-carry and the $0.02$-carry conditions did not train at all in our configuration, so there the carry is a path the optimizer needs, if not for memory. Whether any of this extends to natural language, whose structure is plausibly cross-channel in the learned representation, and to Parcae's RMSNorm, large-scale setting, is a hypothesis this paper does not test.

\section{Limitations}
\label{sec:limitations}

\textbf{Scale.} Training experiments use $d \le 32$ and at most 4 heads, so queries and keys are $8$-dimensional; whether attention at this size computes what transformer attention computes at scale is part of the same untested transfer. Random-weight spectral measurements extend to $d = 256$. The exact results are dimension-independent; quantitative behavior at larger scale, and any trained-model behavior beyond $d = 32$, is untested.

\textbf{Seeds.} The central windowed-copy claims are seed-replicated (five learned-carry, ten zero-carry seeds, each varying initialization and batch stream). The rest of the training evidence is single-run: Tiny Shakespeare, the unconstrained-carry sweep, the intermediate frozen conditions, the post-hoc override, the retention observation, the per-channel task, and the trained-model Jacobians (two inputs of one model). The ratio $C/F$ is seed-replicated at $d = 32$ only (\cref{tab:cf}). Accuracy is measured on $64$ draws with replacement from the $14$ source symbols at each gap, a sampling measurement on a small support rather than generalization to unseen sequences; the curriculum confounds gap with accumulated training.

\textbf{Spectral measurements.} All spectral statements are measurements over finitely many states, seeds, and operating points; no a priori bound on $\|J_f\|_2$ or $\rho_{\text{spec}}(J_G)$ is derived. The natural certifying routes do not deliver one here: by the block-triangular structure the spectrum is determined by the per-token self-blocks $J_{ss}$, but $\max_s \|J_{ss}\|_2 \approx 1.9$--$2.1\,(1-\rho)$ and its log-norm refinement $\max_s \mu_2(J_{ss}) \approx 1.2$--$1.3\,(1-\rho)$ (seed 7, five carries), so both bounds exceed $1$ at every operating point, and the non-normality resides predominantly within tokens. A certificate would need an instance-adapted (Lyapunov-type) norm. The spectral margin is an asymptotic-rate margin; for a non-normal $J_G$ it does not bound how small a perturbation of the weights or the operating point could move an eigenvalue across the unit circle, a pseudospectral question~\citep{trefethen2005} we do not pursue. The non-settling initializations are classified by a residual plateau within $2 \times 10^4$ iterations; this does not prove eventual non-convergence, and their fixed points were not located. On the width ladder the per-instance margins are exact eigenvalues, but the still-decaying seeds were measured away from a fixed point and the non-settling seeds not at all (\cref{tab:width}).

\textbf{Fixed-depth training.} The dispensability result and the depth sweep of \cref{fig:depth-sweep} are fixed-depth results. The sampled-depth replication (\cref{tab:stochastic-depth}) is nine runs over a single sampled range and budget: two learned-carry seeds, one each at $\bar{A} = 0.5$ and $0.02$, four zero-carry seeds, and one learning-rate probe; it shows that the small-carry conditions do not train there, not why, and larger ranges, longer budgets, intermediate carry values, and Parcae's per-sequence sampling are untested.

\textbf{Normalization.} The gain-control coupling follows from homogeneity and carries over to RMSNorm and ScaleNorm. The centering-specific results (the per-token mean-shift kernel of $J_f$ and the exact eigenvalue of \cref{eq:margin-cap}) are LayerNorm-specific; at $\varepsilon = 0$ the RMSNorm Jacobian instead annihilates the radial direction. Parcae's trained models use RMSNorm, and our training experiments use LayerNorm only; we have not run a block without normalization or with RMSNorm. Post-LN architectures are not analyzed.

\textbf{Task boundary.} The dispensability claim is over the tested cross-channel tasks in their given representations; hybrid tasks and larger gaps ($k$ approaching $T$) remain open, and the per-channel contrast is not a controlled isolation of what triggers recruitment. ``Cross-channel'' and ``axis-aligned'' are representation-dependent, since trainable projections can place useful quantities into individual channels. Other carry parameterizations or regularizers than Parcae's were not tested against the unconstrained-carry instability.

\section{Reproducibility}
\label{sec:reproducibility}

All experiments run on CPU. Training runs, power-iteration measurements, and the trained-model Jacobians of \cref{subsec:intervention} use pure Clojure; the direct eigendecomposition/SVD cross-checks, the dense-Jacobian values of \cref{tab:spectral,tab:per-channel}, the eigenvalue verification of \cref{eq:margin-cap}, the width ladder, and the implosion probe use the repository's PyTorch backend (CPU, float64). The codebase implements dense linear algebra, multi-head causal self-attention (full and windowed), pre-LN blocks, the gated recurrence (simplified and faithful modes), manual backpropagation through shared weights, and the training loop from scratch; all gradients are verified against central finite differences (tolerance $10^{-6}$). Total compute is hours on a consumer laptop. Code, with a table-to-script map in its README, is at \ifanon an anonymized repository provided as supplementary material\else\url{https://github.com/clojure-finance/mythjure}, tag \texttt{paper-v2}\fi.

\section{Conclusion}

In a looped transformer with pre-LayerNorm blocks and a diagonal carry, LayerNorm ties the block's sensitivity to the carry: at a fixed point, the increment's Lipschitz constant equals $1 - \rho$ times a ratio that is nearly constant across the scalar carries tested. This shrinks the one-step bound toward $1$ but does not make the loop contractive. With scalar carry, every fixed point we verified is nonetheless locally stable: the Jacobian is strongly non-normal, with operator norm above $1$ but spectral radius below. The spectral margin, the spectral radius's distance from $1$, shrinks as the carry approaches $1$ and, at high carry, with width. The carry constraint guarantees bounded trajectories and a fixed point, not convergence to it.

On the copy tasks tested, the carry is not needed to learn the task at fixed depth. Frozen at $0.5$, $0.02$, or $0$, it trains as well as a learned carry, which changes little during training; a model trained with a learned carry nonetheless depends on it at run time and breaks if it is zeroed. The trained computation is read out in a window of a few iterations around the training depth. With the depth sampled per batch, carries of $0$ and $0.02$ fail to train while $0.5$ succeeds, so there the carry's size matters even though training still changes it little. On a constructed per-channel task, training moves the carry substantially. These results suggest that Parcae's carry constraint costs little on such tasks because learning them at fixed depth does not need the carry; whether this holds at scale and with RMSNorm is untested.

\ifanon\else
\section*{Acknowledgments}

AI language-model assistance was used for mathematical formalization, code generation, and drafting. The author is solely responsible for all claims, experimental design, and interpretation.
\fi

\bibliographystyle{tmlr}
\bibliography{references}

@article{parcae2026,
  author = {Prairie, Hayden and Novack, Zachary and Berg-Kirkpatrick, Taylor and Fu, Daniel Y.},
  title = {Parcae: Scaling Laws For Stable Looped Language Models},
  journal = {arXiv preprint arXiv:2604.12946},
  year = {2026}
}

@inproceedings{s4,
  author = {Gu, Albert and Goel, Karan and R{\'e}, Christopher},
  title = {Efficiently Modeling Long Sequences with Structured State Spaces},
  booktitle = {International Conference on Learning Representations},
  year = {2022}
}

@inproceedings{mamba,
  author = {Gu, Albert and Dao, Tri},
  title = {{Mamba}: Linear-Time Sequence Modeling with Selective State Spaces},
  booktitle = {Conference on Language Modeling (COLM)},
  year = {2024}
}

@inproceedings{mamba2,
  author = {Dao, Tri and Gu, Albert},
  title = {Transformers are {SSMs}: Generalized Models and Efficient Algorithms Through Structured State Space Duality},
  booktitle = {International Conference on Machine Learning},
  year = {2024}
}

@inproceedings{saunshi2025,
  author = {Saunshi, Nikunj and Dikkala, Nishanth and Li, Zhiyuan and Kumar, Sanjiv and Reddi, Sashank J.},
  title = {Reasoning with Latent Thoughts: On the Power of Looped Transformers},
  booktitle = {International Conference on Learning Representations},
  year = {2025}
}

@inproceedings{geiping2025,
  author = {Geiping, Jonas and McLeish, Sean and Jain, Neel and Kirchenbauer, John and Singh, Siddharth and Bartoldson, Brian R. and Kailkhura, Bhavya and Bhatele, Abhinav and Goldstein, Tom},
  title = {Scaling up Test-Time Compute with Latent Reasoning: A Recurrent Depth Approach},
  booktitle = {Advances in Neural Information Processing Systems (NeurIPS)},
  volume = {38},
  year = {2025}
}

@inproceedings{hippo,
  author = {Gu, Albert and Dao, Tri and Ermon, Stefano and Rudra, Atri and R{\'e}, Christopher},
  title = {{HiPPO}: Recurrent Memory with Optimal Polynomial Projections},
  booktitle = {Advances in Neural Information Processing Systems},
  year = {2020}
}

@inproceedings{s4d,
  author = {Gu, Albert and Gupta, Ankit and Goel, Karan and R{\'e}, Christopher},
  title = {On the Parameterization and Initialization of Diagonal State Space Models},
  booktitle = {Advances in Neural Information Processing Systems (NeurIPS)},
  volume = {35},
  year = {2022}
}

@inproceedings{kohli2026,
  author = {Kohli, Harsh and Parthasarathy, Srinivasan and Sun, Huan and Yao, Yuekun},
  title = {Loop, Think, \& Generalize: Implicit Reasoning in Recurrent-Depth Transformers},
  booktitle = {Conference on Language Modeling (COLM)},
  year = {2026},
  note = {arXiv:2604.07822}
}

@book{granas2003,
  author = {Granas, Andrzej and Dugundji, James},
  title = {Fixed Point Theory},
  publisher = {Springer},
  year = {2003}
}

@article{layernorm,
  author = {Ba, Jimmy Lei and Kiros, Jamie Ryan and Hinton, Geoffrey E.},
  title = {Layer Normalization},
  journal = {arXiv preprint arXiv:1607.06450},
  year = {2016}
}

@inproceedings{tomihari2025,
  author = {Tomihari, Akiyoshi and Karakida, Ryo},
  title = {Recurrent Self-Attention Dynamics: An Energy-Agnostic Perspective from {Jacobians}},
  booktitle = {Advances in Neural Information Processing Systems (NeurIPS)},
  year = {2025}
}

@inproceedings{deq,
  title={Deep Equilibrium Models},
  author={Bai, Shaojie and Kolter, J. Zico and Koltun, Vladlen},
  booktitle={Advances in Neural Information Processing Systems (NeurIPS)},
  year={2019}
}

@inproceedings{monotonedeq,
  title={Monotone Operator Equilibrium Networks},
  author={Winston, Ezra and Kolter, J. Zico},
  booktitle={Advances in Neural Information Processing Systems (NeurIPS)},
  year={2020}
}

@inproceedings{arora2019,
  author = {Arora, Sanjeev and Li, Zhiyuan and Lyu, Kaifeng},
  title = {Theoretical Analysis of Auto Rate-Tuning by Batch Normalization},
  booktitle = {International Conference on Learning Representations (ICLR)},
  year = {2019}
}

@inproceedings{xiong2020,
  author = {Xiong, Ruibin and Yang, Yunchang and He, Di and Zheng, Kai and Zheng, Shuxin and Xing, Chen and Zhang, Huishuai and Lan, Yanyan and Wang, Liwei and Liu, Tie-Yan},
  title = {On Layer Normalization in the Transformer Architecture},
  booktitle = {International Conference on Machine Learning (ICML)},
  year = {2020}
}

@inproceedings{kim2021,
  author = {Kim, Hyunjik and Papamakarios, George and Mnih, Andriy},
  title = {The {Lipschitz} Constant of Self-Attention},
  booktitle = {International Conference on Machine Learning (ICML)},
  year = {2021}
}

@book{trefethen2005,
  author = {Trefethen, Lloyd N. and Embree, Mark},
  title = {Spectra and Pseudospectra: The Behavior of Nonnormal Matrices and Operators},
  publisher = {Princeton University Press},
  year = {2005}
}

@inproceedings{kerg2019,
  author = {Kerg, Giancarlo and Goyette, Kyle and Touzel, Maximilian Puelma and Gidel, Gauthier and Vorontsov, Eugene and Bengio, Yoshua and Lajoie, Guillaume},
  title = {Non-normal Recurrent Neural Network ({nnRNN}): Learning Long Time Dependencies While Improving Expressivity with Transient Dynamics},
  booktitle = {Advances in Neural Information Processing Systems (NeurIPS)},
  year = {2019}
}

@inproceedings{orhan2020,
  author = {Orhan, A. Emin and Pitkow, Xaq},
  title = {Improved Memory in Recurrent Neural Networks with Sequential Non-normal Dynamics},
  booktitle = {International Conference on Learning Representations (ICLR)},
  year = {2020}
}

@inproceedings{bai2021,
  author = {Bai, Shaojie and Koltun, Vladlen and Kolter, J. Zico},
  title = {Stabilizing Equilibrium Models by {Jacobian} Regularization},
  booktitle = {International Conference on Machine Learning (ICML)},
  year = {2021}
}

@inproceedings{dehghani2019,
  author = {Dehghani, Mostafa and Gouws, Stephan and Vinyals, Oriol and Uszkoreit, Jakob and Kaiser, {\L}ukasz},
  title = {Universal Transformers},
  booktitle = {International Conference on Learning Representations (ICLR)},
  year = {2019}
}

@inproceedings{giannou2023,
  author = {Giannou, Angeliki and Rajput, Shashank and Sohn, Jy-yong and Lee, Kangwook and Lee, Jason D. and Papailiopoulos, Dimitris},
  title = {Looped Transformers as Programmable Computers},
  booktitle = {International Conference on Machine Learning (ICML)},
  year = {2023}
}

@inproceedings{yang2024,
  author = {Yang, Liu and Lee, Kangwook and Nowak, Robert and Papailiopoulos, Dimitris},
  title = {Looped Transformers are Better at Learning Learning Algorithms},
  booktitle = {International Conference on Learning Representations (ICLR)},
  year = {2024}
}

@inproceedings{orvieto2023,
  title     = {Resurrecting Recurrent Neural Networks for Long Sequences},
  author    = {Orvieto, Antonio and Smith, Samuel L. and Gu, Albert and Fernando, Anushan and Gulcehre, Caglar and Pascanu, Razvan and De, Soham},
  booktitle = {Proceedings of the 40th International Conference on Machine Learning (ICML)},
  year      = {2023}
}

@inproceedings{pascanu2013,
  title     = {On the Difficulty of Training Recurrent Neural Networks},
  author    = {Pascanu, Razvan and Mikolov, Tomas and Bengio, Yoshua},
  booktitle = {Proceedings of the 30th International Conference on Machine Learning (ICML)},
  year      = {2013}
}

@inproceedings{srivastava2015,
  title     = {Training Very Deep Networks},
  author    = {Srivastava, Rupesh Kumar and Greff, Klaus and Schmidhuber, J{\"u}rgen},
  booktitle = {Advances in Neural Information Processing Systems (NeurIPS)},
  volume    = {28},
  year      = {2015}
}

@inproceedings{bansal2022,
  title     = {End-to-end Algorithm Synthesis with Recurrent Networks: Logical Extrapolation Without Overthinking},
  author    = {Bansal, Arpit and Schwarzschild, Avi and Borgnia, Eitan and Emam, Zeyad and Huang, Furong and Goldblum, Micah and Goldstein, Tom},
  booktitle = {Advances in Neural Information Processing Systems (NeurIPS)},
  volume    = {35},
  year      = {2022}
}

@inproceedings{schwarzschild2021,
  title     = {Can You Learn an Algorithm? Generalizing from Easy to Hard Problems with Recurrent Networks},
  author    = {Schwarzschild, Avi and Borgnia, Eitan and Gupta, Arjun and Huang, Furong and Vishkin, Uzi and Goldblum, Micah and Goldstein, Tom},
  booktitle = {Advances in Neural Information Processing Systems (NeurIPS)},
  volume    = {34},
  year      = {2021}
}

@inproceedings{miller2019,
  title     = {Stable Recurrent Models},
  author    = {Miller, John and Hardt, Moritz},
  booktitle = {International Conference on Learning Representations (ICLR)},
  year      = {2019}
}

@inproceedings{noci2022,
  title     = {Signal Propagation in Transformers: Theoretical Perspectives and the Role of Rank Collapse},
  author    = {Noci, Lorenzo and Anagnostidis, Sotiris and Biggio, Luca and Orvieto, Antonio and Singh, Sidak Pal and Lucchi, Aurelien},
  booktitle = {Advances in Neural Information Processing Systems (NeurIPS)},
  volume    = {35},
  year      = {2022}
}

@inproceedings{dong2021,
  title     = {Attention is not all you need: Pure attention loses rank doubly exponentially with depth},
  author    = {Dong, Yihe and Cordonnier, Jean-Baptiste and Loukas, Andreas},
  booktitle = {Proceedings of the 38th International Conference on Machine Learning (ICML)},
  pages     = {2793--2803},
  year      = {2021}
}

@inproceedings{he2016,
  title     = {Deep Residual Learning for Image Recognition},
  author    = {He, Kaiming and Zhang, Xiangyu and Ren, Shaoqing and Sun, Jian},
  booktitle = {Proceedings of the IEEE Conference on Computer Vision and Pattern Recognition (CVPR)},
  pages     = {770--778},
  year      = {2016}
}

\appendix
\crefalias{section}{appendix}
\section{The Inverse-Norm Coupling}
\label{app:coupling}

Consider first a block with a single pre-LN sublayer, whose increment is $u(h) = W_2 \cdot \text{act}(W_1 \cdot \text{LN}(h))$; the two-sublayer composition actually used is treated below. In the gated recurrence the increment $f$ is this term without any identity: the block's internal residual path is folded into $\bar{A} \odot h_t$, so $f(h_t, e) = \text{Block}(h_t + e) - (h_t + e)$ contains no identity in $h_t$. (Were $f$ to include the identity, the recurrence would carry $(\bar{A} + I) \odot h_t$ with diagonal entries exceeding $1$.)

For one token, with centering projection $P = I - \tfrac{1}{d}\mathbf{1}\mathbf{1}^\top$ and spread $s(x) = \sqrt{\tfrac{1}{d}\|Px\|^2 + \varepsilon}$, LayerNorm (dropping $\beta$) is $\text{LN}(x) = \gamma \odot Px/s(x)$, with
\begin{equation}
J_{\text{LN}}(x) = \frac{1}{s(x)}\,\text{diag}(\gamma)\left(P - \frac{1}{d\,s(x)^2}(Px)(Px)^\top\right),
\qquad
\|J_{\text{LN}}(x)\|_2 \le \frac{\|\gamma\|_\infty}{s(x)} .
\label{eq:jln}
\end{equation}

\begin{proof}[Proof of \cref{prop:homogeneity}]
Differentiating $N(\alpha x) = N(x)$ in $x$ gives $\alpha\,J_N(\alpha x) = J_N(x)$; differentiating in $\alpha$ at $\alpha = 1$ gives $J_N(x)\,x = 0$. For LayerNorm, the bracketed matrix in \cref{eq:jln} acts as $0$ on $\mathbf{1}$ (since $P\mathbf{1} = 0$ and $(Px)^\top \mathbf{1} = 0$), as the factor $1 - \|Px\|^2/(d\,s(x)^2) \in [0, 1]$ on $Px$, and as the identity on the rest of the centered subspace. Hence it has spectral norm at most $1$, which gives the bound, and $J_{\text{LN}}(x)\,\mathbf{1} = 0$; $\beta$ does not enter the Jacobian.
\end{proof}

The downstream maps $W_1, W_2$ contribute constant factors, so $\text{Lip}(f) = \|\partial u/\partial h\| = C/\|h\| + o(1/\|h\|)$ under the RMS proxy, where the weight-dependent factor $\|W_1\|\,\|W_2\|\,\|\gamma\|_\infty \sup|\text{act}'|$ is an upper bound on $C$; the measured local $C$ of \cref{tab:cf} additionally depends on the operating-point direction through the softmax and activation derivatives, which is why it is measured rather than derived.

\paragraph{Scope.} The homogeneity part of \cref{prop:homogeneity} applies verbatim to RMSNorm ($P = I$, $s = \text{RMS}$) and to ScaleNorm. For $\varepsilon > 0$ the radial identity acquires the exact correction $J_{\text{LN}}(x)\,x = \varepsilon\,s(x)^{-3}\,\gamma \odot Px$ (verified against finite differences). The shift invariance $J_{\text{LN}}(x)\,\mathbf{1} = 0$, which comes from centering and holds for every $\varepsilon$, is what \cref{prop:mean-shift} uses.

\paragraph{The composed block is not exactly homogeneous.} The increment chains two sublayers with an internal residual, $f(x) = a(x) + m(x + a(x))$ with $a = \text{Attn} \circ \text{LN}_1$ and $m = \text{MLP} \circ \text{LN}_2$. At $\varepsilon = 0$ each sublayer is scale-invariant in its own input, but $f(\alpha x) = a(x) + m(\alpha x + a(x)) \ne f(x)$ in general. The chain rule gives $J_f(x) = J_a(x) + J_m(x + a(x))\,(I + J_a(x))$; where the state dominates the mixture, both terms inherit the inverse scale, and at smaller scales the attention increment's $O(1)$ contribution to $\text{LN}_2$'s input is an additional contributor to the drift of the measured $C$.

\paragraph{At the fixed point.} \Cref{prop:balance} gives the exact scale $\|h^*\| = F/q_D(h^*)$, hence
\begin{equation*}
\text{Lip}(f)\big|_{h^*} = \frac{C}{F}\,q_D(h^*) \;\xrightarrow{\text{scalar carry}}\; \frac{C}{F}\,(1 - \rho),
\end{equation*}
an identity given the definitions of $C$ and $F$, whose affine shape in $\rho$ holds conditionally on $C/F$ being near-constant, which \cref{tab:cf} measures. The scale variable entering the normalizer is $s(h^* + e)$ rather than $\|h^*\|$; measured, $\|h^* + e\|/\|h^*\|$ falls from $1.23$ at $\rho = 0.3$ to $1.002$ at $\rho = 0.95$, so the injection inflates the normalizer's scale by $\approx 23\%$ at low carry, consistent in sign with the lowest $C/F$ occurring there.

\section{Existence of the Fixed Point via Boundedness}
\label{app:existence}

\begin{proof}[Proof of \cref{prop:existence}]
\textbf{Uniform bound on $f$.} Every path from input to output of $f$ passes through a normalizer. Per token, $\|\text{LN}(x)\| \le \|\gamma\|_\infty \|Px\|/s + \|\beta\| \le \|\gamma\|_\infty \sqrt{d} + \|\beta\|$, since $\|Px\|/\sqrt{\|Px\|^2/d + \varepsilon} \le \sqrt{d}$. Downstream, attention outputs are convex combinations of bounded value vectors and the MLP increment is bounded because $|\text{act}(z)| \le |z|$ for the standard activations. Summing the sublayer increments over $n$ tokens gives $\|f(h, e)\| \le B$ with $B$ determined by the weights, $\gamma$, $\beta$, and the dimensions.

\textbf{Trajectories.} Since $D$ scales every channel by at most $\rho$, $\|h_{t+1}\| \le \rho\,\|h_t\| + B$, and induction gives $\|h_t\| \le \rho^t \|h_0\| + B(1 - \rho^t)/(1 - \rho)$.

\textbf{Invariant ball and Brouwer.} Let $r = B/(1 - \rho)$. For $\|h\| \le r$, $\|G(h)\| \le \rho\, r + B = r$, so the map $G$, which is continuous for $\varepsilon > 0$, sends the closed ball $B_r$ into itself and Brouwer's theorem~\citep{granas2003} yields $h^* \in B_r$ with $G(h^*) = h^*$ and $\|h^*\| \le B/(1-\rho)$. A per-channel invariant product of balls with radii $B_i/(1-\bar{A}_i)$ follows the same way. \end{proof}

For the non-settling initializations of \cref{tab:seed-robustness} the fixed point guaranteed here is not reached from the initialization used, and the iteration remains on a bounded orbit inside $B_r$.

\section{Activation Implosion}
\label{sec:implosion}

The inverse coupling creates an asymmetric risk. Large spread lowers $\text{Lip}(f)$, and once $\rho + \text{Lip}(f) < 1$ the operator-norm bound itself certifies contraction. Small spread spikes it: in the rescaling experiment of \cref{fig:implosion-law}, with state and injection imploded together, $\text{Lip}(f)$ reaches $\approx 980$ at a per-token spread of $0.004$ and the full map is locally expansive there, with $\rho_{\text{spec}}(J_G)$ up to $192$.

With fixed weights this does not produce divergence, since trajectories stay bounded by \cref{prop:existence}. Perturbations injected at an imploded state are amplified up to $\times 147$ over the next $T = 8$ iterations (Jacobian products along the trajectory), yet every perturbed trajectory, across both rescaling modes, both carries, and all implosion depths tested, re-converges to the original fixed point; the return to that fixed point is an empirical observation beyond the boundedness guarantee. The exposure is therefore the high-gain transient. The backward pass traverses the same Jacobian products, so during training, where weights do change, gradients can explode from below: the classic recurrent failure of \citet{pascanu2013}, entered through the activation scale. We measure the potential for amplification, not exploding loss gradients, which also depend on direction and readout, and we have no evidence that training steers toward imploded states.

Two further observations bound the risk. Along the tested state-rescaling trajectories, with the injection $e$ at its natural scale, the normalizer input $h + e$ never imploded below the injection's own spread ($\approx 0.8$), capping $\text{Lip}(f) \approx 4$ (\cref{fig:implosion-law}); the full spike required the whole stream, injection included, to implode. This floor is trajectory-relative, not structural: at the constructed state $h = c\mathbf{1} - e$ the first normalizer's input has zero spread while $e$'s does not, and we measure $\text{Lip}(f) \approx 3.8 \times 10^4$ there ($\varepsilon$-capped). Parcae independently reports the complementary failure at 1.3B scale, where growing prelude output norms destabilize the first loop iteration from above~\citep[Fig.~21--22]{parcae2026}; together the two bracket the operating regime through the same scale variable. A practical monitor is the smallest per-token spread at the normalizer inputs, which the global $\|h\|$ only proxies: a single imploded token spikes the block's local Lipschitz constant while the global norm looks healthy.

\section{Unconstrained Carry: Reproducing the Instability}
\label{subsec:unconstrained}

Left unconstrained, does gradient descent drive the carry past $\rho = 1$? We compare the Parcae-constrained carry $\bar{A} = \exp(-\exp(\lambda))$ against an unconstrained scalar carry across learning rates on Tiny Shakespeare ($d=24$, $T=8$, 120 steps; \cref{tab:unconstrained}). The constrained carry stays below $1$ at every rate; the free carry stays below $1$ at low rates but reaches $1.20$ at lr $3 \times 10^{-2}$ and $1.84$ at lr $10^{-1}$, with the state norm growing to $\sim 10^4$. The loss stays finite because the readout normalizes $h_T$ before the unembedding, so loss curves alone can hide this failure; and the state plateaus rather than overflowing because at fixed $T = 8$ the linear path contributes the finite factor $\rho^T$ and in-block LayerNorm bounds each iteration's injection. This is the instability the constraint prevents, reproduced at toy scale.

\begin{table}[ht]
\centering
\begin{tabular}{@{}lcccc@{}}
\toprule
 & lr=$10^{-3}$ & lr=$10^{-2}$ & lr=$3 \times 10^{-2}$ & lr=$10^{-1}$ \\
\midrule
Parcae $\rho(\bar{A})$ & 0.512 & 0.606 & 0.790 & 0.777 \\
Free $\rho(\bar{A})$ & 0.534 & 0.755 & 1.197 & 1.841 \\
Free $\|h_T\|$ & 4.0 & 31 & 325 & 9,570 \\
\bottomrule
\end{tabular}
\caption{An unconstrained carry learns $\rho > 1$ at higher learning rates, with state explosion (cf.\ Parcae's Table 2 and Fig.\ 3). One run per cell.}
\label{tab:unconstrained}
\end{table}

\section{Additional Spectral Measurements}
\label{app:width}

\Cref{tab:spectral} gives the direct spectral values for the seed-7 instance of \cref{tab:cf} plotted in \cref{fig:marginal-stability}.

\begin{table}[ht]
\centering
\begin{tabular}{@{}ccccc@{}}
\toprule
$\rho$ & $\rho_{\text{spec}}(J_G)$ & $\|J_G\|_2$ & $\|J_G\|_2 - \rho_{\text{spec}}$ & margin $1-\rho_{\text{spec}}$ \\
\midrule
0.30 & 0.928 & 1.735 & 0.807 & 0.072 \\
0.50 & 0.901 & 1.481 & 0.580 & 0.099 \\
0.70 & 0.968 & 1.302 & 0.334 & 0.032 \\
0.90 & 0.974 & 1.041 & 0.067 & 0.026 \\
0.95 & 0.986 & 1.018 & 0.032 & 0.014 \\
\bottomrule
\end{tabular}
\caption{Spectral radius vs.\ operator norm of $J_G(h^*)$ for the instance of \cref{tab:cf}, by direct eigendecomposition and SVD of the materialized Jacobian (float64) at verified fixed points.}
\label{tab:spectral}
\end{table}

\Cref{tab:width} gives the per-width numbers behind \cref{fig:width}. The only negative exact margins belong to seeds whose residual is still falling at the iteration cap; their Jacobian is taken away from the fixed point, so these are unfinished settlings, not instabilities. Together with estimator bias they account for the power-iteration estimates that straddled zero at $d \ge 128$ in an earlier run; on converged seeds power iteration is within $0.02$ of the exact value. The margin implied by the mean settle time is $0.7$--$0.9$ of the measured one. The ten seeds of \cref{tab:seed-robustness} are the first ten of the $d = 32$ rung. Each width shares one injection draw across its seeds, and the head count changes with width, so the ladder does not isolate width from head count.

\begin{table}[ht]
\centering
\footnotesize
\begin{tabular}{@{}ccccccc@{}}
\toprule
$d$ (heads) & conv.\ / dec.\ / n.s. & margin (conv.) & $< 0$: conv., dec. & $\|J_G\|_2$ & settle iters & implied \\
\midrule
32 (4)   & 75 / 8 / 17  & $0.0206 \pm 0.0012$ & 0/75, 7/8  & $1.07$ & $1.2$k & $0.018$ \\
64 (8)   & 76 / 4 / 20  & $0.0145 \pm 0.0009$ & 0/76, 0/4  & $1.07$ & $1.7$k & $0.012$ \\
128 (16) & 73 / 8 / 19  & $0.0089 \pm 0.0006$ & 0/73, 5/8  & $1.08$ & $3.3$k & $0.006$ \\
256 (32) & 61 / 10 / 29 & $0.0056 \pm 0.0005$ & 0/61, 6/10 & $1.08$ & $4.8$k & $0.004$ \\
\bottomrule
\end{tabular}
\caption{Width ladder at $\rho = 0.9$, $100$ seeds per width, float64; $\rho_{\text{spec}}$ exact from the per-token blocks, $\|J_G\|_2$ by SVD. Seeds are converged (residual $< 10^{-9}$), still decaying at the $2 \times 10^4$-iteration cap, or non-settling. Margin (mean $\pm$ standard error), operator norm, and settle iterations are over converged seeds; ``$< 0$'' counts negative exact margins among converged and decaying seeds; the implied margin is $\ln(10^{9})$ over the mean settle iterations.}
\label{tab:width}
\end{table}

\section{Training Details and Preliminary Tasks}
\label{app:training}

All windowed-copy runs (\cref{subsec:windowed,subsec:intervention,subsec:sampled-depth}) use $d = 32$, $4$ heads, MLP width $128$, sequence length $12$, window $w = 1$, loop depth $T = 8$, and vocabulary $16$ with $14$ content tokens. Weights are initialized Gaussian with standard deviation $1/\sqrt{d}$ for the block's input projections, $1/\sqrt{2d}$ for its output projections, and $0.02$ for the embeddings (seed $0$ unless stated); the carry is initialized at $\bar{A} = 0.5$ ($\lambda = \ln(-\ln 0.5)$) and frozen, where stated, by zeroing its gradient.

Training is Adam (learning rate $1.5 \times 10^{-3}$, $\beta = (0.9, 0.999)$, no clipping, no weight decay) on cross-entropy at the query position, batches of $4$ fresh examples, through the curriculum $k = 2, 3, 4, 5, 6$ with $300, 200, 200, 250, 400$ steps per stage, each stage warm-started from the previous one with the optimizer state carried over. Accuracy is the argmax at the query over $64$ examples drawn fresh from the task's support, which at a given gap has only $14$ source symbols; sampling noise is zero at $1.00$ and about $\pm 0.06$ at intermediate values. The ablation of \cref{subsec:ablation} uses $d = 24$ and a cold start per condition.

\Cref{tab:windowed-copy} gives the per-stage results of the windowed-copy run of record (\cref{subsec:windowed}).

\begin{table}[ht]
\centering
\begin{tabular}{@{}cccc@{}}
\toprule
$k$ & heuristic $1 - 1/k$ & learned $\max(\bar{A})$ & accuracy \\
\midrule
2 & 0.500 & 0.524 & 0.91 \\
3 & 0.667 & 0.533 & 1.00 \\
4 & 0.750 & 0.518 & 1.00 \\
5 & 0.800 & 0.505 & 1.00 \\
6 & 0.833 & 0.486 & 1.00 \\
\bottomrule
\end{tabular}
\caption{Windowed copy ($w = 1$, $T = 8$): every stage solved, carry not recruited. The learned $\max(\bar{A})$ drifts down from $0.524$ to $0.486$ across the curriculum.}
\label{tab:windowed-copy}
\end{table}

\begin{figure}[ht]
\centering
\resizebox{\textwidth}{!}{%
\begin{tikzpicture}[
    box/.style={draw, minimum width=1.7cm, minimum height=0.7cm, align=center, font=\small},
    arr/.style={->, thick},
    cut/.style={->, red, dashed, thick},
    sum/.style={draw, circle, minimum size=0.55cm, inner sep=0pt},
    io/.style={font=\small},
]
\newcommand{\attnicon}[3]{%
    % #1 x, #2 y (bottom-left of 4x4 grid), #3 full(1)/windowed(0)
    \begin{scope}[shift={(#1,#2)}, scale=0.15]
        \fill[gray!12] (0,0) rectangle (4,4);
        \ifnum#3=1
            \fill[blue!45] (0,3) rectangle (1,4);
            \fill[blue!45] (0,2) rectangle (2,3);
            \fill[blue!45] (0,1) rectangle (3,2);
            \fill[blue!45] (0,0) rectangle (4,1);
        \else
            \fill[blue!45] (0,3) rectangle (1,4);
            \fill[blue!45] (0,2) rectangle (2,3);
            \fill[blue!45] (1,1) rectangle (3,2);
            \fill[blue!45] (2,0) rectangle (4,1);
        \fi
        \draw[gray!55, very thin] (0,0) grid (4,4);
        \draw[gray, thin] (0,0) rectangle (4,4);
    \end{scope}
}
\newcommand{\ablunit}[7]{%
    % #1 xshift, #2 block label, #3 h_t->block live(1)/cut(0), #4 title, #5 attn full(1)/win(0), #6 acc, #7 acc color
    \begin{scope}[xshift=#1 cm]
        \node[io] (h) at (0, 0.3) {$h_t$};
        \node[io] (e) at (0, -1.0) {$e$};
        \node[box, fill=blue!15] (carry) at (2.0, 0.6) {$\bar{A} \odot h_t$};
        \node[box, fill=green!15] (block) at (2.0, -0.7) {#2};
        \node[sum] (s) at (3.7, -0.05) {$+$};
        \node[io] (out) at (5.0, -0.05) {$h_{t+1}$};
        \draw[arr] (h) -- (carry);
        \ifnum#3=1
            \draw[arr] (h) -- (block);
        \else
            \draw[cut] (h) -- (block);
            \node[red, font=\footnotesize\bfseries] at (0.78, -0.05) {$\times$};
        \fi
        \draw[arr] (e) -- (block);
        \draw[arr] (carry) -- (s);
        \draw[arr] (block) -- (s);
        \draw[arr] (s) -- (out);
        \node[font=\small\bfseries] at (2.2, 1.55) {#4};
        \attnicon{1.25}{-2.1}{#5}
        \node[io, anchor=west] at (1.95, -1.8) {Attn: \ifnum#5=1 full\else windowed\fi};
        \node[font=\footnotesize\bfseries, text=#7] at (2.2, -2.5) {Acc: #6};
    \end{scope}
}
\ablunit{0}{$f(h_t, e)$}{1}{A: Full}{0}{1.00}{green!50!black}
\ablunit{6.2}{$f(e)$}{0}{B: Ablated}{0}{0.09}{red}
\ablunit{12.4}{$f(e)$}{0}{C: Ablated}{1}{1.00}{green!50!black}
\end{tikzpicture}%
}
\caption{Ablation conditions. Each sums the linear carry and the nonlinear block into $h_{t+1}$. In A the block reads $h_t$; in B and C that path is cut (red $\times$) and the block sees only the loop-constant input $e$. The grid is the attention mask (banded: windowed $w{=}1$; triangular: full causal). A and C solve; B, ablated and windowed, fails.}
\label{fig:ablation}
\end{figure}

\paragraph{Preliminary tasks.} Two tasks preceded the windowed copy task; neither puts real demand on depth memory.

\paragraph{Tiny Shakespeare.}\label{subsec:tinyshakes} With no engineered memory pressure, does a natural task move the carry? A character-level model ($d=32$, 4 heads, $T=4$, per-channel $\lambda$ initialized at $0.5$, i.e.\ $\bar{A} = 0.192$; the later experiments initialize $\bar{A} = 0.5$, so the levels are not comparable across sections) was trained on Tiny Shakespeare for 150 steps (one run). The loss fell from $4.13$ to $3.05$ (uniform baseline $\ln 65 \approx 4.17$) while $\max(\bar{A})$ drifted from $0.192$ to $0.233$, with the per-channel gradient $\partial L/\partial \bar{A}$ split roughly evenly in sign across channels at every logged step. At $\bar{A} \approx 0.23$ the linear path's e-folding time is $-1/\ln \bar{A} \approx 0.7$ iterations (geometric-sum horizon $\tau \approx 1.3$), so no channel holds long-horizon depth memory. This short run is a weak null: the task gave the optimizer little reason to move the carry, and the question needs a task that demands memory across iterations.

\paragraph{Full-attention copy.}\label{subsec:full-attn} A synthetic copy task, $[A][\text{pad} \times k][?]$ with the loss at the query, demands that a symbol be remembered across a gap. With full causal attention, however, the gap is bridged by attention in a single hop regardless of $k$, so the task puts no demand on depth memory at all. A short sweep over $k \in \{2, 6, 8, 12\}$ (120 steps each, one run, six-example evaluation) accordingly showed no relation between $k$ and the learned carry; its training budget was too small to solve the task reliably, and we draw no further conclusion from it. The carry persists state across loop iterations at a fixed position; it cannot move information across positions. Forcing depth memory requires restricting attention.

\section{Retention of Earlier Gaps and the Post-Hoc Override}
\label{app:retention}

\Cref{tab:retention} records what the final model of each condition retains of the earlier gaps after the sequential curriculum. The nonzero-carry conditions drift off the earlier gaps, a property of warm-starting through stages, whereas the zero-carry model of this run retains almost everything ($0.91$--$1.00$ across $k = 2$--$6$). The direction replicates across the training seeds of \cref{subsec:intervention}, the size does not: on the earliest gap $k = 2$ the nine solving zero-carry seeds retain $0.39$--$1.00$ (median $0.63$) against $0.00$--$0.23$ for the five learned-carry seeds, and on $k = 5$ they retain $0.78$--$1.00$ against $0.16$--$0.64$. Each checkpoint solves its current stage in every condition. We do not have an explanation for the difference.

\begin{table}[ht]
\centering
\begin{tabular}{@{}lccccc@{}}
\toprule
Final model & $k=2$ & $k=3$ & $k=4$ & $k=5$ & $k=6$ \\
\midrule
(a) learned carry & 0.16 & 0.08 & 0.13 & 0.48 & 1.00 \\
(b) frozen $\bar{A} = 0.5$ & 0.06 & 0.05 & 0.02 & 0.66 & 1.00 \\
(c) frozen $\bar{A} = 0.02$ & 0.14 & 0.41 & 0.39 & 0.84 & 1.00 \\
(d) frozen $\bar{A} = 0$ & 0.91 & 0.92 & 1.00 & 1.00 & 1.00 \\
\midrule
(a) with $\bar{A} := 0$ post hoc & 0.13 & 0.11 & 0.11 & 0.16 & 0.14 \\
(a) with $\bar{A} := 0.5$ post hoc & 0.06 & 0.00 & 0.13 & 0.61 & 1.00 \\
\bottomrule
\end{tabular}
\caption{Accuracy of each condition's final model (after the $k = 6$ stage) on every gap, same protocol as \cref{tab:intervention}; chance $\approx 0.07$. Top: retention of earlier gaps. Bottom: post-hoc carry override of the learned-carry model without retraining.}
\label{tab:retention}
\end{table}

\section{A Minimal Contraction Check}
\label{app:toy}

Before the full block, the contraction bookkeeping was checked on the recurrence with $f(h, e) = \tanh(\text{gain} \cdot Wh + e)$ for a random $8 \times 8$ matrix $W$ at $\rho = 0.368$ (no attention, no LayerNorm). The contraction-mapping (Banach) condition $\rho + \text{gain} \cdot \|W\|_2 < 1$ with the measured $\|W\|_2 \approx 5.39$ guarantees convergence for gain $< 0.117$; empirically the loop settles for gain $\le 0.25$ and wanders without settling for gain $\ge 0.30$, because the bound charges $\tanh$ its global slope of $1$ while the local slope at the fixed point is smaller. The sufficient-not-necessary structure recurs at full block scale in \cref{sec:self-stabilization}. The approach to the fixed point curves and rotates (\cref{fig:spiral}): step sizes decay overall with a small rebound around $t \approx 8$ while successive update directions rotate, the signature of complex Jacobian eigenstructure. The rebound is suggestive of non-normal transient amplification but is not measured on this system; the non-normality of the full block is measured in \cref{subsec:directional}. \Cref{fig:three-fates} uses the same construction with an independent draw of $W$.

\begin{figure}[t]
\centering
\begin{minipage}{0.44\textwidth}
\centering
\begin{tikzpicture}
\begin{axis}[
    width=\textwidth,
    height=\textwidth,
    axis equal image,
    title={(a) trajectory (PCA projection)},
    xlabel={PC 1}, ylabel={PC 2},
    grid=major, grid style={gray!25},
    tick label style={font=\footnotesize},
    label style={font=\footnotesize},
    title style={font=\footnotesize},
]
% 2D PCA projection of the d=8 nonlinear loop (gain 0.25), centered on the fixed point.
\addplot[blue, thick, mark=*, mark size=1.1pt, mark options={fill=blue}] coordinates {
    (-2.3978,-1.6901) (-2.1683,0.6430) (-1.3083,1.2066) (-0.7023,1.0064)
    (-0.3563,0.5972) (-0.1396,0.2207) (-0.0098,-0.0505) (0.0587,-0.2093)
    (0.0896,-0.2765) (0.0979,-0.2804) (0.0903,-0.2451) (0.0704,-0.1884)
    (0.0414,-0.1217) (0.0081,-0.0536) (-0.0233,0.0087) (-0.0468,0.0586)
    (-0.0586,0.0905) (-0.0580,0.1017) (-0.0477,0.0935) (-0.0319,0.0711)
    (-0.0152,0.0418) (-0.0006,0.0127) (0.0099,-0.0109) (0.0157,-0.0262)
    (0.0172,-0.0327) (0.0152,-0.0315) (0.0107,-0.0246) (0.0050,-0.0144)
};
\addplot[red, mark=star, mark size=3pt, only marks] coordinates {(0,0)};
\node[anchor=west, font=\scriptsize] at (axis cs:-2.35,-1.35) {$h_0$};
\node[anchor=west, font=\scriptsize, red] at (axis cs:0.10,0.06) {$h^*$};
\end{axis}
\end{tikzpicture}
\end{minipage}\hfill
\begin{minipage}{0.54\textwidth}
\centering
\begin{tikzpicture}
\begin{axis}[
    width=\textwidth,
    height=0.85\textwidth,
    title={(b) step size and direction vs.\ iteration},
    xlabel={iteration $t$},
    xmin=0, xmax=15,
    ymin=-0.75, ymax=1.1,
    legend pos=north east,
    legend style={font=\scriptsize, fill opacity=0.85, text opacity=1},
    grid=major, grid style={gray!25},
    tick label style={font=\footnotesize},
    label style={font=\footnotesize},
    title style={font=\footnotesize},
]
\addplot[black, dotted, sharp plot, mark=none, forget plot] coordinates {(0,0) (15,0)};
% step size normalized by the first step (2.437): shrinking
\addplot[blue, thick, mark=o, mark size=1.6pt] coordinates {
    (0,1.000) (1,0.461) (2,0.296) (3,0.252) (4,0.208) (5,0.139) (6,0.083)
    (7,0.065) (8,0.072) (9,0.078) (10,0.075) (11,0.066) (12,0.053)
    (13,0.040) (14,0.029) (15,0.025)
};
\addlegendentry{$\|\Delta h_t\| / \|\Delta h_0\|$}
% cosine of each update vs the first: rotating (sweeps to $-0.63$ and back)
\addplot[red, thick, mark=square, mark size=1.4pt] coordinates {
    (0,1.000) (1,0.444) (2,-0.291) (3,-0.574) (4,-0.598) (5,-0.629) (6,-0.611)
    (7,-0.310) (8,0.009) (9,0.165) (10,0.251) (11,0.319) (12,0.393)
    (13,0.475) (14,0.517) (15,0.390)
};
\addlegendentry{$\cos(\Delta h_t, \Delta h_0)$}
\end{axis}
\end{tikzpicture}
\end{minipage}
\caption{Rotating approach of the settling loop ($d=8$ toy, gain $0.25$, $\rho=0.368$). \textbf{(a)} Trajectory projected onto its top two principal components, approaching $h^*$ along a curved path. \textbf{(b)} Step size (blue) decays with a small rebound near $t \approx 8$ while the update direction (red) rotates, its cosine against the first update sweeping from $1$ through $0$ to $-0.63$ and back: complex Jacobian eigenstructure.}
\label{fig:spiral}
\end{figure}

\begin{figure}[ht]
\centering
\begin{tikzpicture}
\begin{semilogyaxis}[
    width=0.8\textwidth,
    height=0.48\textwidth,
    xlabel={loop iteration $t$},
    ylabel={step size $\|h_{t+1} - h_t\|$},
    xmin=0, xmax=40,
    ymin=1e-8, ymax=1e4,
    legend pos=south west,
    legend style={font=\footnotesize, fill opacity=0.9, text opacity=1},
    grid=major, grid style={gray!25},
    tick label style={font=\footnotesize},
    label style={font=\footnotesize},
]
\addplot[blue, thick, mark=o, mark size=1.1pt, mark repeat=2] coordinates {
(0,2.041e+00) (1,1.159e+00) (2,4.922e-01) (3,2.295e-01) (4,1.344e-01) (5,8.342e-02) (6,5.051e-02) (7,3.076e-02) (8,1.959e-02) (9,1.297e-02) (10,8.661e-03) (11,5.759e-03) (12,3.823e-03) (13,2.552e-03) (14,1.720e-03) (15,1.168e-03) (16,7.978e-04) (17,5.463e-04) (18,3.746e-04) (19,2.570e-04) (20,1.765e-04) (21,1.213e-04) (22,8.332e-05) (23,5.725e-05) (24,3.934e-05) (25,2.703e-05) (26,1.857e-05) (27,1.276e-05) (28,8.762e-06) (29,6.019e-06) (30,4.134e-06) (31,2.840e-06) (32,1.950e-06) (33,1.340e-06) (34,9.202e-07) (35,6.320e-07) (36,4.341e-07) (37,2.982e-07) (38,2.048e-07) (39,1.407e-07) (40,9.661e-08)
};
\addlegendentry{settles (locally stable)}
\addplot[orange!90!black, thick, mark=square, mark size=1.1pt, mark repeat=2] coordinates {
(0,2.228e+00) (1,1.746e+00) (2,1.019e+00) (3,1.130e+00) (4,1.211e+00) (5,1.253e+00) (6,1.016e+00) (7,1.129e+00) (8,1.183e+00) (9,1.210e+00) (10,1.051e+00) (11,1.133e+00) (12,1.169e+00) (13,1.177e+00) (14,1.051e+00) (15,1.123e+00) (16,1.160e+00) (17,1.169e+00) (18,1.049e+00) (19,1.120e+00) (20,1.158e+00) (21,1.168e+00) (22,1.048e+00) (23,1.119e+00) (24,1.158e+00) (25,1.168e+00) (26,1.048e+00) (27,1.119e+00) (28,1.158e+00) (29,1.168e+00) (30,1.048e+00) (31,1.119e+00) (32,1.158e+00) (33,1.168e+00) (34,1.048e+00) (35,1.119e+00) (36,1.158e+00) (37,1.168e+00) (38,1.048e+00) (39,1.119e+00) (40,1.158e+00)
};
\addlegendentry{bounded, non-settling}
\addplot[red!80!black, thick, mark=triangle, mark size=1.3pt, mark repeat=2] coordinates {
(0,1.948e+00) (1,2.847e+00) (2,3.641e+00) (3,4.453e+00) (4,5.369e+00) (5,6.447e+00) (6,7.736e+00) (7,9.282e+00) (8,1.114e+01) (9,1.336e+01) (10,1.602e+01) (11,1.921e+01) (12,2.303e+01) (13,2.757e+01) (14,3.296e+01) (15,3.940e+01) (16,4.717e+01) (17,5.657e+01) (18,6.787e+01) (19,8.145e+01) (20,9.774e+01) (21,1.173e+02) (22,1.407e+02) (23,1.689e+02) (24,2.027e+02) (25,2.432e+02) (26,2.918e+02) (27,3.502e+02) (28,4.202e+02) (29,5.043e+02) (30,6.052e+02) (31,7.262e+02) (32,8.714e+02) (33,1.046e+03) (34,1.255e+03) (35,1.506e+03) (36,1.807e+03) (37,2.168e+03) (38,2.602e+03) (39,3.122e+03) (40,3.747e+03)
};
\addlegendentry{grows (carry unconstrained)}
\end{semilogyaxis}
\end{tikzpicture}
\caption{Three regimes of an iterated recurrence, on the $d=8$ tanh toy of \cref{app:toy}. Blue (carry $0.368$, gain $0.25$): the step size decays geometrically and the state settles. Orange (same carry, gain $0.8$): the step size neither decays nor grows; the state circulates on a bounded orbit, as some random initializations of the full block do (\cref{tab:seed-robustness}). Red (gain $0.25$, carry $1.2$): the state diverges.}
\label{fig:three-fates}
\end{figure}

\end{document}